\documentclass[11pt]{article}
\usepackage[margin=1in]{geometry}
\usepackage{amsmath}
\usepackage{booktabs}
\usepackage{xcolor}
\usepackage{longtable}
\usepackage{array}
\usepackage{graphicx}
\usepackage{hyperref}
\usepackage{parskip}
\usepackage{natbib}
\usepackage[T1]{fontenc}
\hypersetup{colorlinks=true, linkcolor=black, urlcolor=blue}
\usepackage{setspace}
\newcommand{\inputtable}[1]{\IfFileExists{#1}{\input{#1}}{}}
\newcommand{\JevChoiceBias}{\ensuremath{-}0.0373}

\newcommand{\JevFlipRate}{8.8}
\newcommand{\JevProbAsym}{\ensuremath{-}0.0358}

\newcommand{\JevDecisivePairs}{27,367}
\newcommand{\JevFlipRateTwo}{8.5}
\newcommand{\JevProbAsymTwo}{0.0030}

\newcommand{\JevTransFwd}{98.18}

\newcommand{\PaperTransLo}{95.95}
\newcommand{\PaperTransHi}{99.06}

\newcommand{\JevZeta}{0.9700}
\newcommand{\JevCircular}{639}
\newcommand{\JevTriads}{82,160}
\newcommand{\JevRRItems}{80}
\newcommand{\JevWeakStoch}{98.37}
\newcommand{\JevStrongStoch}{80.21}
\newcommand{\JevStochTriples}{11,841}

\newcommand{\JevFonePrimary}{0.814}

\newcommand{\JevFoneCILo}{0.764}
\newcommand{\JevFoneCIHi}{0.852}

\newcommand{\HumanNPairs}{243}

\newcommand{\JevFoneGapTop}{0.026}
\newcommand{\JevFoneGapTopLo}{\ensuremath{-}0.001}
\newcommand{\JevFoneGapTopHi}{0.059}
\newcommand{\FoneCIWidthLo}{0.08}
\newcommand{\FoneCIWidthHi}{0.13}

\newcommand{\JevFwdRevR}{0.982}

\newcommand{\JevUncertEffect}{0.256}

\newcommand{\JevUncertScoreEffect}{\ensuremath{-}0.190}
\newcommand{\JevUncertScoreLo}{\ensuremath{-}0.313}
\newcommand{\JevUncertScoreHi}{\ensuremath{-}0.067}
\newcommand{\JevControlEffect}{0.093}
\newcommand{\JevControlLo}{\ensuremath{-}0.034}
\newcommand{\JevControlHi}{0.221}

\newcommand{\PaperControlNPos}{2}
\newcommand{\PaperControlNNeg}{1}
\newcommand{\RetestIdentical}{98.6}
\newcommand{\RetestMeanDp}{0.0119}
\newcommand{\RetestR}{0.998}
\newcommand{\RetestN}{493}
\newcommand{\RetestItemR}{0.999}

\providecommand{\GilNTwentyOne}{}\renewcommand{\GilNTwentyOne}{2,403}
\providecommand{\GilNTwentyThree}{}\renewcommand{\GilNTwentyThree}{411}
\providecommand{\GilAccOne}{}\renewcommand{\GilAccOne}{89.0}
\providecommand{\GilAccPen}{}\renewcommand{\GilAccPen}{88.2}
\providecommand{\GilAccOneTwentyThree}{}\renewcommand{\GilAccOneTwentyThree}{76.2}
\providecommand{\GilAccPenTwentyThree}{}\renewcommand{\GilAccPenTwentyThree}{74.2}
\providecommand{\GilUnanimous}{}\renewcommand{\GilUnanimous}{97.9}
\providecommand{\GilPenaltyCost}{}\renewcommand{\GilPenaltyCost}{0.8}
\providecommand{\GilProblemAcc}{}\renewcommand{\GilProblemAcc}{92.4}
\providecommand{\GilSolutionAcc}{}\renewcommand{\GilSolutionAcc}{84.7}
\providecommand{\GilProblemN}{}\renewcommand{\GilProblemN}{2,361}
\providecommand{\GilSolutionN}{}\renewcommand{\GilSolutionN}{2,398}
\providecommand{\GilIca}{}\renewcommand{\GilIca}{98.7}

\providecommand{\GilChatGPTOne}{}\renewcommand{\GilChatGPTOne}{72.8}
\providecommand{\GilChatGPTTwo}{}\renewcommand{\GilChatGPTTwo}{70.2}
\providecommand{\GilMturk}{}\renewcommand{\GilMturk}{71.2}
\providecommand{\GilHumanIca}{}\renewcommand{\GilHumanIca}{93.9}
\providecommand{\GilChatGPTIcaOne}{}\renewcommand{\GilChatGPTIcaOne}{92.0}
\providecommand{\GilChatGPTIcaTwo}{}\renewcommand{\GilChatGPTIcaTwo}{95.0}

\providecommand{\GilMargin}{}\renewcommand{\GilMargin}{15.5}

\providecommand{\GilRequests}{}\renewcommand{\GilRequests}{12,156}
\providecommand{\GilCost}{}\renewcommand{\GilCost}{0.61}

\providecommand{\ParN}{}\renewcommand{\ParN}{907}
\providecommand{\ParNTotal}{}\renewcommand{\ParNTotal}{945}

\providecommand{\ParGptFour}{}\renewcommand{\ParGptFour}{0.789}
\providecommand{\ParJevFew}{}\renewcommand{\ParJevFew}{0.740}

\providecommand{\ParTweetNLP}{}\renewcommand{\ParTweetNLP}{0.659}

\providecommand{\ParGapFourFew}{}\renewcommand{\ParGapFourFew}{0.049}
\providecommand{\ParGapFourFewLo}{}\renewcommand{\ParGapFourFewLo}{0.016}
\providecommand{\ParGapFourFewHi}{}\renewcommand{\ParGapFourFewHi}{0.084}
\providecommand{\ParGapNlpFew}{}\renewcommand{\ParGapNlpFew}{0.081}
\providecommand{\ParGapNlpFewLo}{}\renewcommand{\ParGapNlpFewLo}{0.030}
\providecommand{\ParGapNlpFewHi}{}\renewcommand{\ParGapNlpFewHi}{0.132}
\providecommand{\ParGapNlpPcaFew}{}\renewcommand{\ParGapNlpPcaFew}{0.055}
\providecommand{\ParGapNlpPcaFewLo}{}\renewcommand{\ParGapNlpPcaFewLo}{0.006}
\providecommand{\ParGapNlpPcaFewHi}{}\renewcommand{\ParGapNlpPcaFewHi}{0.106}
\providecommand{\ParGapNlpZero}{}\renewcommand{\ParGapNlpZero}{0.042}
\providecommand{\ParGapNlpZeroLo}{}\renewcommand{\ParGapNlpZeroLo}{\ensuremath{-}0.010}
\providecommand{\ParGapNlpZeroHi}{}\renewcommand{\ParGapNlpZeroHi}{0.098}
\providecommand{\ParGapNlpPcaZero}{}\renewcommand{\ParGapNlpPcaZero}{0.055}
\providecommand{\ParGapNlpPcaZeroLo}{}\renewcommand{\ParGapNlpPcaZeroLo}{0.002}
\providecommand{\ParGapNlpPcaZeroHi}{}\renewcommand{\ParGapNlpPcaZeroHi}{0.107}

\providecommand{\ParBootB}{}\renewcommand{\ParBootB}{2,000}

\providecommand{\DilChesLo}{}\renewcommand{\DilChesLo}{0.856}
\providecommand{\DilChesHi}{}\renewcommand{\DilChesHi}{0.891}
\providecommand{\DilChesMean}{}\renewcommand{\DilChesMean}{0.877}
\providecommand{\DilChesTheirMean}{}\renewcommand{\DilChesTheirMean}{0.772}

\providecommand{\DilChesN}{}\renewcommand{\DilChesN}{8}

\providecommand{\DilYears}{}\renewcommand{\DilYears}{9}
\providecommand{\DilYearLo}{}\renewcommand{\DilYearLo}{1979}
\providecommand{\DilYearHi}{}\renewcommand{\DilYearHi}{2019}
\providecommand{\DilFirstShare}{}\renewcommand{\DilFirstShare}{46.6}

\providecommand{\DilTevNLo}{}\renewcommand{\DilTevNLo}{17}
\providecommand{\DilTevNHi}{}\renewcommand{\DilTevNHi}{93}

\providecommand{\FxOutTokDileo}{}\renewcommand{\FxOutTokDileo}{38}

\providecommand{\FxOutTokGilardi}{}\renewcommand{\FxOutTokGilardi}{97}

\providecommand{\FxOutTokSchemas}{}\renewcommand{\FxOutTokSchemas}{6}

\providecommand{\FxProbZeroShare}{}\renewcommand{\FxProbZeroShare}{14.8}

\providecommand{\FxSumOneShare}{}\renewcommand{\FxSumOneShare}{99.86}

\providecommand{\FxSumShortN}{}\renewcommand{\FxSumShortN}{721}

\providecommand{\FxConfFormula}{}\renewcommand{\FxConfFormula}{\ensuremath{(K\,p_{\max}-1)/(K-1)}}

\providecommand{\FxConfPreN}{}\renewcommand{\FxConfPreN}{1,567}
\providecommand{\FxConfPreWithin}{}\renewcommand{\FxConfPreWithin}{100.0}
\providecommand{\FxConfPostWithin}{}\renewcommand{\FxConfPostWithin}{86.7}

\providecommand{\FxChoiceN}{}\renewcommand{\FxChoiceN}{517,039}
\providecommand{\FxArgmaxShare}{}\renewcommand{\FxArgmaxShare}{99.987}

\providecommand{\FxScoreN}{}\renewcommand{\FxScoreN}{6,432}

\providecommand{\FxScoreFeasible}{}\renewcommand{\FxScoreFeasible}{99.8}

\providecommand{\FxRetestChoiceN}{}\renewcommand{\FxRetestChoiceN}{1,000}

\providecommand{\FxRetestSameChoice}{}\renewcommand{\FxRetestSameChoice}{98.0}
\providecommand{\FxRetestSameVector}{}\renewcommand{\FxRetestSameVector}{40.7}
\providecommand{\FxRetestMeanAbsDiff}{}\renewcommand{\FxRetestMeanAbsDiff}{0.0120}

\providecommand{\FxRetestGapMinutes}{}\renewcommand{\FxRetestGapMinutes}{47}

\providecommand{\FxDupMeanAbsDiff}{}\renewcommand{\FxDupMeanAbsDiff}{0.0061}

\providecommand{\FxBundleN}{}\renewcommand{\FxBundleN}{150}

\providecommand{\FxBundleMeanAbsDp}{}\renewcommand{\FxBundleMeanAbsDp}{0.0085}

\providecommand{\FxTotalOk}{}\renewcommand{\FxTotalOk}{495,943}

\providecommand{\FxDateFirst}{}\renewcommand{\FxDateFirst}{19 September 2026}
\providecommand{\FxDateLast}{}\renewcommand{\FxDateLast}{22 September 2026}

\providecommand{\BrtJevBareAcc}{}\renewcommand{\BrtJevBareAcc}{0.715}
\providecommand{\BrtJevBareFone}{}\renewcommand{\BrtJevBareFone}{0.555}
\providecommand{\BrtJevBareKnownFone}{}\renewcommand{\BrtJevBareKnownFone}{0.644}

\providecommand{\BrtJevBareUnkRecall}{}\renewcommand{\BrtJevBareUnkRecall}{0.002}
\providecommand{\BrtJevCbAcc}{}\renewcommand{\BrtJevCbAcc}{0.721}
\providecommand{\BrtJevCbFone}{}\renewcommand{\BrtJevCbFone}{0.638}
\providecommand{\BrtJevCbKnownFone}{}\renewcommand{\BrtJevCbKnownFone}{0.734}

\providecommand{\BrtJevCbUnkRecall}{}\renewcommand{\BrtJevCbUnkRecall}{0.009}
\providecommand{\BrtQwenBareAcc}{}\renewcommand{\BrtQwenBareAcc}{0.654}
\providecommand{\BrtQwenBareFone}{}\renewcommand{\BrtQwenBareFone}{0.435}
\providecommand{\BrtQwenBareKnownFone}{}\renewcommand{\BrtQwenBareKnownFone}{0.503}

\providecommand{\BrtQwenBareUnkRecall}{}\renewcommand{\BrtQwenBareUnkRecall}{0.028}
\providecommand{\BrtQwenCbAcc}{}\renewcommand{\BrtQwenCbAcc}{0.636}
\providecommand{\BrtQwenCbFone}{}\renewcommand{\BrtQwenCbFone}{0.559}
\providecommand{\BrtQwenCbKnownFone}{}\renewcommand{\BrtQwenCbKnownFone}{0.634}

\providecommand{\BrtQwenCbUnkRecall}{}\renewcommand{\BrtQwenCbUnkRecall}{0.072}
\providecommand{\BrtCbertAcc}{}\renewcommand{\BrtCbertAcc}{0.840}
\providecommand{\BrtCbertFone}{}\renewcommand{\BrtCbertFone}{0.744}
\providecommand{\BrtCbertKnownFone}{}\renewcommand{\BrtCbertKnownFone}{0.775}

\providecommand{\BrtCbertUnkRecall}{}\renewcommand{\BrtCbertUnkRecall}{0.550}
\providecommand{\BrtCllQeightAcc}{}\renewcommand{\BrtCllQeightAcc}{0.765}
\providecommand{\BrtCllQeightFone}{}\renewcommand{\BrtCllQeightFone}{0.647}
\providecommand{\BrtCllQeightKnownFone}{}\renewcommand{\BrtCllQeightKnownFone}{0.687}

\providecommand{\BrtCllQeightUnkRecall}{}\renewcommand{\BrtCllQeightUnkRecall}{0.387}
\providecommand{\BrtCllQfourAcc}{}\renewcommand{\BrtCllQfourAcc}{0.729}
\providecommand{\BrtCllQfourFone}{}\renewcommand{\BrtCllQfourFone}{0.571}
\providecommand{\BrtCllQfourKnownFone}{}\renewcommand{\BrtCllQfourKnownFone}{0.608}

\providecommand{\BrtCllQfourUnkRecall}{}\renewcommand{\BrtCllQfourUnkRecall}{0.323}
\providecommand{\BrtQwenTwoAcc}{}\renewcommand{\BrtQwenTwoAcc}{0.541}
\providecommand{\BrtQwenTwoFone}{}\renewcommand{\BrtQwenTwoFone}{0.397}
\providecommand{\BrtQwenTwoKnownFone}{}\renewcommand{\BrtQwenTwoKnownFone}{0.457}

\providecommand{\BrtQwenTwoUnkRecall}{}\renewcommand{\BrtQwenTwoUnkRecall}{0.060}
\providecommand{\BrtLlamaAcc}{}\renewcommand{\BrtLlamaAcc}{0.346}
\providecommand{\BrtLlamaFone}{}\renewcommand{\BrtLlamaFone}{0.208}
\providecommand{\BrtLlamaKnownFone}{}\renewcommand{\BrtLlamaKnownFone}{0.236}

\providecommand{\BrtLlamaUnkRecall}{}\renewcommand{\BrtLlamaUnkRecall}{0.072}
\providecommand{\BrtGemmaAcc}{}\renewcommand{\BrtGemmaAcc}{0.325}
\providecommand{\BrtGemmaFone}{}\renewcommand{\BrtGemmaFone}{0.214}
\providecommand{\BrtGemmaKnownFone}{}\renewcommand{\BrtGemmaKnownFone}{0.230}

\providecommand{\BrtGemmaUnkRecall}{}\renewcommand{\BrtGemmaUnkRecall}{0.280}
\providecommand{\BrtLunaBareAcc}{}\renewcommand{\BrtLunaBareAcc}{0.709}
\providecommand{\BrtLunaBareFone}{}\renewcommand{\BrtLunaBareFone}{0.544}
\providecommand{\BrtLunaBareKnownFone}{}\renewcommand{\BrtLunaBareKnownFone}{0.631}

\providecommand{\BrtLunaBareUnkRecall}{}\renewcommand{\BrtLunaBareUnkRecall}{0.001}
\providecommand{\BrtLunaCbAcc}{}\renewcommand{\BrtLunaCbAcc}{0.743}
\providecommand{\BrtLunaCbFone}{}\renewcommand{\BrtLunaCbFone}{0.634}
\providecommand{\BrtLunaCbKnownFone}{}\renewcommand{\BrtLunaCbKnownFone}{0.725}

\providecommand{\BrtLunaCbUnkRecall}{}\renewcommand{\BrtLunaCbUnkRecall}{0.029}
\providecommand{\BrtN}{}\renewcommand{\BrtN}{37,709}
\providecommand{\BrtNKnown}{}\renewcommand{\BrtNKnown}{33,381}
\providecommand{\BrtNUnknown}{}\renewcommand{\BrtNUnknown}{4,328}

\providecommand{\BrtJevBarePredUnk}{}\renewcommand{\BrtJevBarePredUnk}{0.17}
\providecommand{\BrtJevCbPredUnk}{}\renewcommand{\BrtJevCbPredUnk}{0.95}

\providecommand{\BrtJevBareAssRecall}{}\renewcommand{\BrtJevBareAssRecall}{0.285}
\providecommand{\BrtJevCbAssRecall}{}\renewcommand{\BrtJevCbAssRecall}{0.542}

\providecommand{\BrtQwenBareAssRecall}{}\renewcommand{\BrtQwenBareAssRecall}{0.015}

\providecommand{\BrtOvrAcc}{}\renewcommand{\BrtOvrAcc}{0.640}
\providecommand{\BrtOvrFone}{}\renewcommand{\BrtOvrFone}{0.413}

\providecommand{\BrtOvrTies}{}\renewcommand{\BrtOvrTies}{380}
\providecommand{\BrtChoiceTies}{}\renewcommand{\BrtChoiceTies}{44}
\providecommand{\BrtOvrOnlyChoice}{}\renewcommand{\BrtOvrOnlyChoice}{3,534}
\providecommand{\BrtOvrOnlyOvr}{}\renewcommand{\BrtOvrOnlyOvr}{728}

\providecommand{\BrtOvrKnownFone}{}\renewcommand{\BrtOvrKnownFone}{0.478}
\providecommand{\BrtOvrUnkRecall}{}\renewcommand{\BrtOvrUnkRecall}{0.001}
\providecommand{\BrtMlChoiceSubset}{}\renewcommand{\BrtMlChoiceSubset}{66.6}
\providecommand{\BrtMlOvrSubset}{}\renewcommand{\BrtMlOvrSubset}{24.9}
\providecommand{\BrtMlOvrCard}{}\renewcommand{\BrtMlOvrCard}{1.76}
\providecommand{\BrtMlTrueCard}{}\renewcommand{\BrtMlTrueCard}{0.96}
\providecommand{\BrtMlCbertSubset}{}\renewcommand{\BrtMlCbertSubset}{79.4}
\providecommand{\BrtMlQeightSubset}{}\renewcommand{\BrtMlQeightSubset}{72.4}
\providecommand{\BrtBbcN}{}\renewcommand{\BrtBbcN}{322}
\providecommand{\BrtBbcJevAcc}{}\renewcommand{\BrtBbcJevAcc}{0.829}
\providecommand{\BrtBbcJevFone}{}\renewcommand{\BrtBbcJevFone}{0.321}
\providecommand{\BrtBbcCbertFone}{}\renewcommand{\BrtBbcCbertFone}{0.681}
\providecommand{\BrtBbcLlamaFone}{}\renewcommand{\BrtBbcLlamaFone}{0.322}
\providecommand{\BrtBbcGemmaFone}{}\renewcommand{\BrtBbcGemmaFone}{0.288}
\providecommand{\BrtGateCells}{}\renewcommand{\BrtGateCells}{71}

\providecommand{\EaGilRelFone}{}\renewcommand{\EaGilRelFone}{0.874}
\providecommand{\EaGilRelRec}{}\renewcommand{\EaGilRelRec}{0.824}
\providecommand{\EaGilRelPrec}{}\renewcommand{\EaGilRelPrec}{0.931}
\providecommand{\EaGilRelPos}{}\renewcommand{\EaGilRelPos}{46.2}
\providecommand{\EaGilRelMaj}{}\renewcommand{\EaGilRelMaj}{53.8}

\providecommand{\EaGilProbFone}{}\renewcommand{\EaGilProbFone}{0.806}

\providecommand{\EaGilProbPos}{}\renewcommand{\EaGilProbPos}{16.7}
\providecommand{\EaGilProbMaj}{}\renewcommand{\EaGilProbMaj}{83.3}

\providecommand{\EaGilProbRecTwentyThree}{}\renewcommand{\EaGilProbRecTwentyThree}{1.000}
\providecommand{\EaGilProbPrecTwentyThree}{}\renewcommand{\EaGilProbPrecTwentyThree}{0.072}
\providecommand{\EaGilProbPosTwentyThree}{}\renewcommand{\EaGilProbPosTwentyThree}{3.0}
\providecommand{\EaGilProbMajTwentyThree}{}\renewcommand{\EaGilProbMajTwentyThree}{97.0}
\providecommand{\EaGilProbAccTwentyThree}{}\renewcommand{\EaGilProbAccTwentyThree}{61.4}
\providecommand{\EaGilProbNTwentyThree}{}\renewcommand{\EaGilProbNTwentyThree}{166}
\providecommand{\EaGilSolFone}{}\renewcommand{\EaGilSolFone}{0.546}
\providecommand{\EaGilSolRec}{}\renewcommand{\EaGilSolRec}{0.408}

\providecommand{\EaGilSolPos}{}\renewcommand{\EaGilSolPos}{22.5}
\providecommand{\EaGilSolMaj}{}\renewcommand{\EaGilSolMaj}{77.5}

\providecommand{\HcNMulti}{}\renewcommand{\HcNMulti}{140}
\providecommand{\HcNJudgMulti}{}\renewcommand{\HcNJudgMulti}{357}
\providecommand{\HcAgree}{}\renewcommand{\HcAgree}{75.2}
\providecommand{\HcCeiling}{}\renewcommand{\HcCeiling}{87.1}
\providecommand{\HcJevAcc}{}\renewcommand{\HcJevAcc}{80.1}

\providecommand{\EaGtdMultiN}{}\renewcommand{\EaGtdMultiN}{2,709}
\providecommand{\EaGtdMultiShare}{}\renewcommand{\EaGtdMultiShare}{7.2}

\providecommand{\DilCovD}{}\renewcommand{\DilCovD}{0.149}
\providecommand{\DilCovDLo}{}\renewcommand{\DilCovDLo}{0.036}
\providecommand{\DilCovDHi}{}\renewcommand{\DilCovDHi}{0.255}
\providecommand{\DilCovRCov}{}\renewcommand{\DilCovRCov}{0.618}
\providecommand{\DilCovRCovLo}{}\renewcommand{\DilCovRCovLo}{0.525}
\providecommand{\DilCovRCovHi}{}\renewcommand{\DilCovRCovHi}{0.699}
\providecommand{\DilCovRUnc}{}\renewcommand{\DilCovRUnc}{0.469}
\providecommand{\DilCovRUncLo}{}\renewcommand{\DilCovRUncLo}{0.393}
\providecommand{\DilCovRUncHi}{}\renewcommand{\DilCovRUncHi}{0.542}
\providecommand{\DilCovNPos}{}\renewcommand{\DilCovNPos}{6}
\providecommand{\DilCovNYears}{}\renewcommand{\DilCovNYears}{8}

\providecommand{\DilCovPYUnc}{}\renewcommand{\DilCovPYUnc}{966}
\providecommand{\DilCovPYUncCountry}{}\renewcommand{\DilCovPYUncCountry}{774}

\providecommand{\DilCovSdCov}{}\renewcommand{\DilCovSdCov}{20.1}
\providecommand{\DilCovSdUnc}{}\renewcommand{\DilCovSdUnc}{19.8}

\providecommand{\LuChesMean}{}\renewcommand{\LuChesMean}{0.844}

\providecommand{\LuChesN}{}\renewcommand{\LuChesN}{8}

\providecommand{\LuRelTwentyOne}{}\renewcommand{\LuRelTwentyOne}{91.1}
\providecommand{\LuRelTwentyOneJev}{}\renewcommand{\LuRelTwentyOneJev}{89.0}

\providecommand{\LuRelTwentyThree}{}\renewcommand{\LuRelTwentyThree}{84.2}
\providecommand{\LuRelTwentyThreeJev}{}\renewcommand{\LuRelTwentyThreeJev}{76.2}

\providecommand{\LuParFew}{}\renewcommand{\LuParFew}{0.816}

\providecommand{\LuParGap}{}\renewcommand{\LuParGap}{0.076}
\providecommand{\LuParGapLo}{}\renewcommand{\LuParGapLo}{0.039}
\providecommand{\LuParGapHi}{}\renewcommand{\LuParGapHi}{0.112}

\providecommand{\LuBrtBareAcc}{}\renewcommand{\LuBrtBareAcc}{0.709}
\providecommand{\LuBrtBareFone}{}\renewcommand{\LuBrtBareFone}{0.544}
\providecommand{\LuBrtCbAcc}{}\renewcommand{\LuBrtCbAcc}{0.743}
\providecommand{\LuBrtCbFone}{}\renewcommand{\LuBrtCbFone}{0.634}

\providecommand{\LuCostRatioLo}{}\renewcommand{\LuCostRatioLo}{0.45}
\providecommand{\LuCostRatioHi}{}\renewcommand{\LuCostRatioHi}{1.03}
\providecommand{\LuCostRatioCwHi}{}\renewcommand{\LuCostRatioCwHi}{1.27}
\providecommand{\LuCostTasks}{}\renewcommand{\LuCostTasks}{10}
\providecommand{\LuRequests}{}\renewcommand{\LuRequests}{510{,}681}
\providecommand{\LuCost}{}\renewcommand{\LuCost}{7.17}
\providecommand{\LuRepeatLo}{}\renewcommand{\LuRepeatLo}{95.3}
\providecommand{\LuRepeatHi}{}\renewcommand{\LuRepeatHi}{97.1}

\providecommand{\LmgEconJev}{}\renewcommand{\LmgEconJev}{0.97}

\providecommand{\LmgEconGpt}{}\renewcommand{\LmgEconGpt}{0.98}

\providecommand{\LmgSocJev}{}\renewcommand{\LmgSocJev}{0.92}

\providecommand{\LmgSocGpt}{}\renewcommand{\LmgSocGpt}{0.94}

\providecommand{\LmgEuJev}{}\renewcommand{\LmgEuJev}{0.81}

\providecommand{\LmgEuGpt}{}\renewcommand{\LmgEuGpt}{0.93}

\providecommand{\LmgEuN}{}\renewcommand{\LmgEuN}{35}

\providecommand{\LmgUkSent}{}\renewcommand{\LmgUkSent}{18,263}
\providecommand{\LmgEuSent}{}\renewcommand{\LmgEuSent}{429}
\providecommand{\LmgUkDocs}{}\renewcommand{\LmgUkDocs}{18}
\providecommand{\LmgEuDocs}{}\renewcommand{\LmgEuDocs}{36}

\providecommand{\VdCountries}{}\renewcommand{\VdCountries}{171}
\providecommand{\VdIndicators}{}\renewcommand{\VdIndicators}{53}

\providecommand{\VdJevCor}{}\renewcommand{\VdJevCor}{0.68}

\providecommand{\VdGptCor}{}\renewcommand{\VdGptCor}{0.63}
\providecommand{\VdLlamaCor}{}\renewcommand{\VdLlamaCor}{0.50}

\providecommand{\VdLunaCor}{}\renewcommand{\VdLunaCor}{0.63}

\providecommand{\VdLunaExact}{}\renewcommand{\VdLunaExact}{52}

\providecommand{\VdLunaPmodeMean}{}\renewcommand{\VdLunaPmodeMean}{0.91}
\providecommand{\VdLunaOverNinetyNine}{}\renewcommand{\VdLunaOverNinetyNine}{52}

\providecommand{\VdLunaCorGap}{}\renewcommand{\VdLunaCorGap}{0.05}
\providecommand{\VdLunaCorGapLo}{}\renewcommand{\VdLunaCorGapLo}{0.04}
\providecommand{\VdLunaCorGapHi}{}\renewcommand{\VdLunaCorGapHi}{0.07}
\providecommand{\VdLunaExactGap}{}\renewcommand{\VdLunaExactGap}{2.7}
\providecommand{\VdLunaExactGapLo}{}\renewcommand{\VdLunaExactGapLo}{1.6}
\providecommand{\VdLunaExactGapHi}{}\renewcommand{\VdLunaExactGapHi}{3.8}
\providecommand{\VdLunaEceGap}{}\renewcommand{\VdLunaEceGap}{0.33}
\providecommand{\VdLunaEceGapLo}{}\renewcommand{\VdLunaEceGapLo}{0.32}
\providecommand{\VdLunaEceGapHi}{}\renewcommand{\VdLunaEceGapHi}{0.35}
\providecommand{\VdLunaEceVal}{}\renewcommand{\VdLunaEceVal}{0.39}

\providecommand{\VdBootDraws}{}\renewcommand{\VdBootDraws}{2,000}

\providecommand{\GptFoFone}{}\renewcommand{\GptFoFone}{0.828}

\providecommand{\GptFoFoneGap}{}\renewcommand{\GptFoFoneGap}{0.014}
\providecommand{\GptFoFoneGapLo}{}\renewcommand{\GptFoFoneGapLo}{\ensuremath{-0.014}}
\providecommand{\GptFoFoneGapHi}{}\renewcommand{\GptFoFoneGapHi}{0.043}
\providecommand{\GptFoSTwo}{}\renewcommand{\GptFoSTwo}{0.296}
\providecommand{\GptFoSTwoLo}{}\renewcommand{\GptFoSTwoLo}{0.166}
\providecommand{\GptFoSTwoHi}{}\renewcommand{\GptFoSTwoHi}{0.425}
\providecommand{\JevSTwo}{}\renewcommand{\JevSTwo}{0.256}
\providecommand{\JevSTwoLo}{}\renewcommand{\JevSTwoLo}{0.130}
\providecommand{\JevSTwoHi}{}\renewcommand{\JevSTwoHi}{0.383}

\title{JEV versus LLMs: Accuracy, Cost and Calibration on Seven Political Science Replications}
\author{Steven Denney\thanks{Leiden Institute for Area Studies, Leiden University. \texttt{s.c.denney@hum.leidenuniv.nl}} \and Matthew DiGiuseppe\thanks{Institute of Political Science, Leiden University. \texttt{m.r.di.giuseppe@fsw.leidenuniv.nl}}}
\date{}

\begin{document}
\maketitle

\begin{abstract}
\noindent Large language models (LLMs) annotate and scale political text or constructs by generating text tokens. A new class of models, which TypeSafe markets as ``System One'' models, instead returns decisions and probability distributions across a user-supplied fixed answer set. A commercial model, JEV, is advertised as having a dramatic cost and speed advantage over traditional LLMs along with better calibrated decisions. As such, it might be useful for social scientists looking to quickly and cost-effectively annotate or scale large corpora of text and have a reliable indicator of a classifier's uncertainty. Yet, the accuracy of these claims and the broader model accuracy in social science text-based tasks are not yet established. In this paper, we do just that and hope to establish the suitability of JEV for social science tasks. We compare JEV with LLMs and human coders from published research, and with a current mid-tier commercial LLM (GPT-6 Luna) and an open-weight alternative (Qwen3.8-27B). We find that JEV matches, or comes close to, the capabilities of both LLMs in a variety of tasks. However, we find no cost advantage over GPT-6 Luna at OpenAI's batch prices. Further, we find that, when each question is asked once, JEV's probabilities are better calibrated than GPT-6 Luna's token probabilities, but not consistently better than Qwen3.8-27B's. We conclude that unless researchers have a need for speed, JEV's only obvious advantage is ease of parsing the underlying choice probabilities. 

\vspace{6pt}
\end{abstract}

\clearpage
\doublespacing

\noindent LLMs have democratised text-as-data tools in the social sciences. Prior to LLMs, most scholars had to invest substantial time in learning text-as-data techniques or have the resources to hire crowd-sourced workers \citep{benoit2016} or research assistants to code a training dataset. Now scholars can use R packages \citep{maerz2026quallmer} to process text data with common statistical software and a shallow learning curve. Still, LLMs are not costless. For large datasets and for frontier capabilities, LLMs do carry a substantial (but declining) token cost. Further, scaling tasks that rely on pairwise comparisons across a corpus require about 20 times as many calls as rating each text once and can quickly get expensive \citep{digiuseppe2026,licht2025}.

A new class of models, which TypeSafe markets as ``System One'' models, is marketed as a cheaper and quicker alternative that could further democratise text analysis in the social sciences. The name borrows from the distinction between fast, intuitive ``System 1'' and slow, deliberate ``System 2'' thinking popularised by \citet{kahneman2011}. The analogy implies that the model makes fast, intuitive decisions rather than slow, well-reasoned ones. TypeSafe presents its JEV (1.13), released in September 2026, as its first public model in this class, and says it works differently from a generative LLM. A generative model answers by writing text one token at a time, and each token needs another pass through the model. The researcher then reads a label or a number out of what the model wrote. JEV returns no text. The user sends a piece of text, which TypeSafe calls ``the state'', and one or more questions with a fixed set of answers. According to TypeSafe, the model reads the state once and judges every question against it in parallel, returning all of its answers in one response \citep{typesafe2026}. TypeSafe has not published the architecture, so we cannot check this account (Appendix E). JEV offers three output types. They are a choice from a user-provided list of categories (Choice), an item on a scale with worded levels (Score), and the probability of a yes answer to a yes/no question (Noul). For all output types, the model reports the probability it attaches to each potential option. Some LLM APIs also report token probabilities, the probability the model gave to each candidate token at each position of its reply, which can serve as a measure of uncertainty. Most report only the 5 to 20 most probable candidates at each position, and some report none. Further, it is not clear that these probabilities are informative and well calibrated, since post-training with human feedback tends to make them overconfident \citep{openai2023,tian2023}.  

TypeSafe charges only for input tokens and output is free. Output tokens cost more per token than input tokens, and reasoning models can write hundreds or thousands of them before they answer, so TypeSafe advertises large cost and speed advantages \citep{typesafe2026}. For classification with short labels, however, most of the bill is input. In our tasks GPT-6 Luna wrote about four tokens per answer, while JEV counts the question and every answer option as input and billed 1.2 to 3.3 times as many input tokens per decision as Luna. At the time of writing, the details of JEV's architecture are not public. As such, researchers cannot verify why the model is cheap and fast or what that implies for its output.

Despite this opacity, the model's efficiency potentially holds many benefits for researchers analysing text, or wishing to exploit the knowledge embedded in such models. Yet, we have little existing evidence that JEV or other ``System One'' models are suitable for social science applications. In this paper, we ask whether the cost and speed advantages of JEV (1.13) over LLMs require a sacrifice in performance. Further, we examine TypeSafe's claim of improved calibration, which it attributes to its training method, reinforcement learning for calibrated decisions (RLCD). We replicate seven published studies that evaluated LLMs for annotation, scaling or data generation. We score JEV against each study's human benchmark and compare it with the published models, with GPT-6 Luna (a current low-cost commercial LLM) and with Qwen3.8-27B (an open-weight alternative).

Our analysis finds that JEV performs on par with its contemporary commercial and open-weight alternatives when using human coding as a benchmark. Further, we find little cost advantage compared to (a) GPT-6 Luna called with batch processing or (b) open-weight models run locally on commercial hardware. JEV does, however, have a clear speed advantage. We also find that JEV's probabilities are better calibrated than the token probabilities GPT-6 Luna returns when each question is asked once, on all eight tasks where we compare them. Post-training for chat tends to make LLMs' answer probabilities overconfident \citep{openai2023,tian2023}, and reasoning models often state high confidence in answers that are wrong \citep{mei2025}. GPT-6 Luna fits this pattern: 74\% of its probabilities on the party comparisons are below 0.01 or above 0.99, and OpenAI returns none when it reasons.\footnote{With open weights, token probabilities can still be computed when a model reasons.} Qwen3.8-27B does not (16\%), and its calibration is similar to JEV's when human coding is the benchmark. 

In all, we find that in the analysis of social science text, JEV has only minor advantages over the existing options we tested. It is clearly faster and it has bounded choice probabilities that are easy to parse. 

\section*{What JEV returns}
\label{sec:interface}

A JEV request contains a state and typed questions. The ``state'' is the text to be judged, while each question specifies what the model should decide and supplies the answer options against which it should evaluate that text \citep{typesafe2026}. Users can pick one of three output types: Choice, Score or Noul. ``Choice'' returns a named option and probabilities over user-supplied options. ``Score'' accepts two to ten worded levels and returns their probability-weighted mean, with the probability assigned to each level. ``Noul'' returns the probability of a positive answer to a binary question. Researchers therefore receive an answer distribution directly, without parsing a generated explanation or extracting a category from free text. Beyond these probabilities, a Choice answer also carries a ``confidence'' value. In our outputs it is almost exactly a rescaling of the probability of the chosen option (correlation 0.9997), so we treat it as adding nothing beyond the probabilities and do not analyse it separately.

These probabilities resemble the token probabilities (log probabilities) that many LLM APIs report for their output. In this case, however, the probabilities are defined over the answer options rather than over possible tokens. A log probability describes the next token, so the researcher must map tokens to answer labels, discard mass on tokens that are not labels, and renormalise. Closed APIs also truncate what they report, or report no probabilities at all. For GPT-6 Luna, OpenAI returns at most the five most probable tokens at each position, so on a nine-point scale some levels receive no probability at all, and it returns none when the model reasons before answering. Anthropic does not report token probabilities at all. Even open-weight models accessed through an inference service will often limit the available probabilities. JEV returns a probability for every option the researcher supplies, rounded to two decimal places, regardless of the number of options. 

TypeSafe calls its post-training method reinforcement learning for calibrated decisions (RLCD), which it says optimizes for ``epistemically honest probabilities'', so that higher confidence means higher accuracy \citep{typesafe2026}. A calibrated model's answers given probability 0.8 should be correct about 80\% of the time across many answers. TypeSafe does not publish the training data, the reward or the architecture, so we cannot check the method directly. We can, however, check its implication against ``gold standard'' human benchmarks. Whatever calibration RLCD achieves, it achieves on TypeSafe's training data. Calibration depends on the data as well as the model, so probabilities calibrated on one set of texts need not be calibrated on another. As such, it is no obvious that the RLCD would translate to honest calibration across all uses cases. %

Still, the underlying choice probabilities produced by JEV are potentially its most important contribution. As we mentioned, LLM token probabilities are limited and not always available. Research suggests that the token probabilities of chat-tuned LLMs are often a poor guide to how often an answer is ``right''. The largest pretrained models are reasonably well calibrated on multiple-choice questions whose options are shown to them \citep{kadavath2022}. However, post-training (reinforcement learning) for chat undoes much of this, although a simple temperature adjustment largely restores it \citep{kadavath2022}. On a subset of the MMLU benchmark, GPT-4's expected calibration error on its answer probabilities rose from 0.007 before post-training to 0.074 after post-training \citep{openai2023}. For GPT-3.5-turbo, GPT-4 and Claude 1 and 2, answer probabilities estimated from repeated samples are poorly calibrated, and the confidence a model states when asked is often better calibrated, in some cases halving the calibration error \citep{tian2023}. One proposed contributor is that the reward models used in this training favor responses that state high confidence regardless of their quality \citep{leng2025}. Reasoning models, whose token probabilities commercial APIs usually do not return, often state confidence above 85\% even on answers they get wrong \citep{mei2025}. Below, we compare JEV with GPT-6 Luna and Qwen3.8-27B to see whether its probabilities track correctness better than theirs do.

JEV's exact architecture also remains unknown. TypeSafe states that JEV reads the state once and evaluates the questions in parallel, producing its outputs in one query, but provides no technical account from which we can verify how those evaluations are performed \citep{typesafe2026}. 

\section*{How does JEV compare to LLMs and humans?}

There have been a variety of papers published in political science and the social sciences more broadly to assess the ability of LLMs to complete tasks assigned to human coders and encoder-only models like BERT. Here we replicate several of these analyses, using the validation benchmarks of these existing studies, to see how JEV compares with these benchmarks, previous-generation LLMs, and the two contemporary models we mentioned above. Table~\ref{tab:replications} identifies the papers and tasks within those papers that we replicate here with JEV called via OpenRouter (for V-Dem, TypeSafe's own endpoint), GPT-6 Luna called directly from OpenAI servers, and Qwen3.8-27B called via a third-party inference provider (Together.ai). 

Our replications include five reading (scaling and annotation) applications and two tasks which rely on the models' embedded knowledge to generate new data. Regarding the latter, the first attempts to create an ideological scaling of European political parties \citep{dileo2025} and the second reproduces V-Dem's expert codes for 53 indicators in 2023 \citep{weidmann2026}. We retain each published task and prompt wording where the interface permits, removing reply-format instructions that JEV cannot follow. Appendix A records the prompts, and Appendices C, F and G document changes to answer options and rating scales. 

\begin{table}[htbp]
\centering

\footnotesize
\setlength{\tabcolsep}{1.8pt}
\renewcommand{\arraystretch}{0.97}
\begin{tabular}{>{\raggedright\arraybackslash}p{1.7cm}
                >{\raggedright\arraybackslash}p{2.4cm}
                >{\raggedright\arraybackslash}p{2.8cm}
                >{\raggedright\arraybackslash}p{2.4cm}
                >{\raggedright\arraybackslash}p{3.5cm}
                >{\raggedright\arraybackslash}p{1.3cm}
                >{\raggedright\arraybackslash}p{1.3cm}}
\toprule
 & & & & \multicolumn{3}{l}{JEV compared with} \\ \cmidrule(l){5-7}
Study & Data & Test replicated & Human reference & published models & GPT-6 Luna & Qwen 27B \\
\midrule
\multicolumn{7}{l}{\emph{Annotation}} \\
\citet{gilardi2023} & Tweets on content moderation, 2021 and 2023 & Relevance (relevant or not), four runs; accuracy & Trained annotators; Mechanical Turk workers as a comparator & Better than ChatGPT & Worse & Slightly better (2021); worse (2023)$^{a}$ \\
\citet{ornstein2025} & \ParNTotal{} tweets on two Supreme Court decisions & Sentiment (positive, neutral, negative); correlation of the probability-based score & Mean of three expert ratings & Worse than GPT-3 and GPT-4; better than Twitter RoBERTa and Naive Bayes & Worse & Worse \\
\citet{brandt2026} & \BrtN{} Global Terrorism Database incidents; \BrtBbcN{} BBC news articles & Attack type, nine categories, names only; accuracy and macro F1. BBC news: conflict F1 & GTD coders' attack type & Better than prompted LLMs; worse than fine-tuned ConfliBERT and ConflLlama. BBC: matches Llama 3.1, worse than ConfliBERT & Better acc.; slightly better F1. BBC: better & Better acc.\ and F1. BBC: better \\
\citet{weidmann2026} & \VdIndicators{} V-Dem indicators, \VdCountries{} countries, 2023 (recall) & Ordinal indicator coded from the country name; correlation and exact agreement & V-Dem expert codes & Better than GPT-4o and Llama-3.1 70B & Better & Better \\
\addlinespace
\multicolumn{7}{l}{\emph{Scaling}} \\
\citet{dileo2025} & European parties in \DilYears{} EP election years, \DilYearLo{}--\DilYearHi{} (recall) & Pairwise ``more right-wing'' from party name; Bradley-Terry scale & CHES expert placements; CMP manifesto codes; TEV voter placements & Better than GPT-3.5 & Slightly better & Better \\
\citet{digiuseppe2026} & 1,402 answers on interest rates (Study 1); debt-ceiling answers (Study 2) & Pairwise knowledge and uncertainty; scale and expert-pair F1 & Expert raters' pairwise judgments & Within the range of six published and nine further generative scorers (mid-table) & Slightly worse & Slightly better \\
\citet{lemens2025} & \LmgUkDocs{} UK manifestos; \LmgEuDocs{} EP speeches on coal subsidies & Sentence ratings averaged to documents; correlation & Expert coding (manifestos); crowd coding (speeches) & Within the range of the published models on manifestos; worse than GPT-4o on speeches & Worse & Slightly worse \\
\bottomrule
\end{tabular}
\normalsize
\caption{The seven published applications re-run with JEV. The human reference is what every model is scored against; designs are unchanged except as described in the text. Reading tasks give the model a text, recall tasks only a name. The last three columns compare JEV with the published models on each study's headline measure, and with GPT-6 Luna and the open-weight Qwen3.8-27B (reasoning off) run under the same prompt (paired bootstrap, \citealt{efron1979}; details in Tables~\ref{tab:gpt6luna} and~\ref{tab:qwen27b}). Against GPT-6 Luna and Qwen, better or worse means the paired 95\% interval excludes zero; slightly better or worse means JEV's estimate is higher or lower but the interval includes zero. Against the published models, which have no paired intervals, better or worse means above or below their reported figures, and within range means between the lowest and highest. $^{a}$~The interval's lower bound is 0.2 points and depends on the bootstrap seed. BBC conflict F1: JEV 0.321, Llama 3.1 0.322, ConfliBERT 0.681.}
\label{tab:replications}
\end{table}

We first assess the speed and cost of JEV given that these are the advantages highlighted in its marketing. Table~\ref{tab:cost_speed} compares cost and speed on the primary task of each application. JEV is not cheaper than GPT-6 Luna at OpenAI's Batch prices. Luna costs less on six of the seven tasks and about the same (3\% more) on the seventh. Against Luna's Standard prices, which apply when answers are needed at once rather than within a day, JEV is cheaper on six of the seven tasks. We ran inference on Qwen through an external provider's API (OpenRouter, served by a single host, Parasail, at 8-bit precision).\footnote{Batch processing was not available for this model.} In this case, Qwen was more expensive than both GPT-6 Luna and JEV, at 2.35 to 5.41 times JEV's cost per decision. However, given the model's small size, many scholars can run it on a local commercial machine or better yet on university supercomputer infrastructure at no cost to the researcher.

JEV is faster than the two alternatives. Sending one request at a time, it returned each decision in 0.09 to 0.27 seconds, against 0.75 to 1.04 seconds for Luna and 0.39 to 0.87 for Qwen.\footnote{These timings come from one location on two consecutive days, and throughput with many requests at once depends on the route and rate limits in force, so the table shows the size of the speed difference in our runs rather than a fixed property of any of the models.} We now proceed to examine the accuracy of JEV in each of the annotation and scaling tasks. 
\begin{table}[htbp]
\centering

\footnotesize
\setlength{\tabcolsep}{2.6pt}
\begin{tabular}{l rrrr rrr rrr}
\toprule
 & \multicolumn{4}{c}{\$ per 1,000 decisions} & \multicolumn{3}{c}{Seconds per decision} & \multicolumn{3}{c}{Decisions per second} \\
\cmidrule(lr){2-5}\cmidrule(lr){6-8}\cmidrule(lr){9-11}
Application & JEV & Luna (Batch) & Luna (Std.) & Qwen & JEV & Luna & Qwen & JEV & Luna & Qwen \\
\midrule
\citet{gilardi2023} & 0.017 & 0.017 & 0.034 & 0.085 & 0.09 & 0.77 & 0.44 & 156 & 18 & 31 \\
\citet{ornstein2025} & 0.024 & 0.012 & 0.025 & 0.128 & 0.26 & 0.75 & 0.39 & 47 & 18 & 30 \\
\citet{brandt2026} & 0.027 & 0.018 & 0.035 & 0.139 & 0.27 & 1.04 & 0.87 & 51 & 14 & 14 \\
\citet{weidmann2026}$^{a}$ & 0.020 & 0.012 & 0.024 & 0.058 & 0.26 & 0.75 & 0.41 & 45 & 17 & 31 \\
\citet{dileo2025} & 0.018 & 0.008 & 0.016 & 0.042 & 0.25 & 0.78 & 0.43 & 46 & 20 & 28 \\
\citet{digiuseppe2026} & 0.012 & 0.012 & 0.024 & 0.058 & 0.13 & 0.78 & 0.44 & 93 & 13 & 32 \\
\citet{lemens2025} & 0.014 & 0.009 & 0.019 & 0.058 & 0.13 & 0.87 & 0.49 & 95 & 17 & 21 \\
\bottomrule
\end{tabular}
\normalsize
\caption{Cost and speed of JEV, GPT-6 Luna and Qwen3.8-27B on the primary task of each application. A decision is one answer to one question about one item. Cost: dollars per 1,000 decisions; JEV and Qwen as billed, GPT-6 Luna at OpenAI's posted Batch and Standard rates. Qwen ran through OpenRouter on one host (Parasail, FP8) with no batch rate, so its cost is that of this route, not of running the open weights oneself. Speed: median seconds per decision with one request at a time, and decisions per second with 16 at a time, from 200 timed items per application sent from the Netherlands on 29--30 September 2026. Throughput depends on the route and rate limits and is not a property of the models. $^{a}$~Costs for JEV and Luna are from the run's totals.}
\label{tab:cost_speed}
\end{table}

\subsection*{Tweet relevance}
\label{sec:gilardi}

\citet{gilardi2023} were among the first to show that LLMs could provide high-quality annotations of a common task: annotating tweets. They use tweets from \citet{alizadeh2022content}, who collected English-language tweets about content moderation through Twitter's academic API from January 2020 to April 2021, about 2.6 million tweets after filtering. \citet{gilardi2023} drew a random sample from this corpus and added a new sample of tweets from January 2023, collected the same way.\footnote{Two trained research assistants coded each tweet for relevance and for whether it presents content moderation as a problem or a solution. In their replication archive, both agreed on 2,403 of the 2,559 double-coded tweets in the 2021 sample (93.9\% agreement) and on 411 of 480 in the 2023 sample (85.6\%); we score only these. For the frames, the agreed sets for the two frame questions hold 2,361 and 2,398 tweets in 2021 and 166 and 170 in 2023.} The tasks we replicate are whether a tweet is relevant to content moderation and, separately, whether it frames content moderation as a problem or as a solution. 

JEV beats the GPT-3.5 model used in the original study if we treat the trained annotators as the proper baseline. However, JEV is below GPT-6 Luna in both samples, and level with or below Qwen3.8-27B. GPT-6 Luna reaches \LuRelTwentyOne\% accuracy against JEV's \LuRelTwentyOneJev\% on the 2021 tweets and \LuRelTwentyThree\% against \LuRelTwentyThreeJev\% on the 2023 tweets, with both paired intervals excluding zero.  Qwen3.8-27B (Table~\ref{tab:qwen27b}) is level with JEV on the 2021 tweets (accuracy 88.2\%, a difference of $-0.8$ points [$-2.1$, 0.4]), but above JEV on the 2023 tweets (accuracy 79.6\%, a difference of $+3.4$ [0.2, 6.8]), although this lead depends on the bootstrap seed. Qwen is below GPT-6 Luna on both tweet samples.  On the two frame questions, for which we rebuilt the instructions, Qwen is above JEV on the problem frame for the 2023 tweets ($+7.8$ points [1.2, 14.5]), but below JEV on the solution frame for the 2021 tweets ($-3.4$ [$-4.4$, $-2.3$]). Qwen is level with JEV in the other two cells (Appendix C).
The criterion is the label agreed by two trained research assistants. Table~\ref{tab:gilardi} and Appendix C retain the historical comparison with crowd workers and ChatGPT, along with JEV's repeated-run results.

\subsection*{Tweet sentiment}
\label{sec:ornstein}

\citet{ornstein2025} classify the sentiment in \ParNTotal{} tweets about two Supreme Court decisions and correlate a continuous score with the mean of three expert judgments.\footnote{The tweets are the authors' own collection: posts referring to the US Supreme Court within 24 hours of two decisions, \textit{Masterpiece Cakeshop v.\ Colorado Civil Rights Commission} in 2018 (423 tweets) and \textit{Trump v.\ Mazars} and \textit{Trump v.\ Vance}, released on the same day in 2020 (522 tweets). The three expert coders labelled each tweet independently as positive, negative or neutral (Fleiss' $\kappa = 0.72$) and recoded disagreements in a second round. The benchmark is their mean on a seven-point scale from $-1$ to $+1$. We compare models on the 907 tweets for which every measure is available.} With their six few-shot examples and positive-minus-negative scoring (Table~\ref{tab:parrot}), JEV reaches \ParJevFew{}, above the Twitter-sentiment RoBERTa model (\ParTweetNLP{}) and below GPT-4 (\ParGptFour{}) in correlation with the experts' mean sentiment score. However, both contemporary models agree more closely with the experts than JEV does. GPT-6 Luna, given the same examples and scored by the GPT-4 rule, reaches \LuParFew{}, exceeding JEV by \LuParGap{} [\LuParGapLo{}, \LuParGapHi{}].  Using the examples and scoring the output the same way, Qwen3.8-27B reaches 0.841 [0.811, 0.868], higher than JEV by 0.102 [0.071, 0.131] and level with GPT-6 Luna ($+0.025$ [$-0.001$, 0.052]). Without the examples Qwen3.8-27B answered in full sentences to almost all tweets (941 of 945), so we have no comparable score for the zero-shot prompt. 

\subsection*{Terrorist attack type}
\label{sec:brandt}

\citet{brandt2026} test two classification tasks that are meant to demonstrate test the utility of LLMs for building event datasets. First they classify the attack type of \BrtN{} Global Terrorism Database \citep{lafree2007,start2022gtd} incidents into nine categories using each incident's summary and motive. Next, they use 322 BBC news articles from 2004--05 \citep{greene2006} and label them as conflict-related or not.\footnote{START researchers code each incident from media reports, screened by automated filters, following the GTD codebook. Each incident has a short narrative summary of who did what, when, where and how, an optional free-text motive, and up to three attack types. \citet{brandt2026} train on incidents up to 2016 and test on 37,709 incidents from 2017 to 2020 (11,220, 9,722, 8,457 and 8,310 a year). The model reads the summary with the motive appended and is scored against the first attack type. In the BBC task 53 items are about conflict.} In their study, they compare fine-tuned models (ConfliBERT and ConflLlama) with prompted LLMs (Gemma 2, Llama 3.1 and Qwen 2.5) that receive no codebook, only the category names. We test their original prompting strategy and also test whether providing a codebook, for context, leads to better outcomes.  With category names alone (Table~\ref{tab:brandt}), JEV reaches accuracy \BrtJevBareAcc{} and macro F1 \BrtJevBareFone{} (the unweighted mean of the nine categories' F1 scores; \citealt{sokolova2009}), while GPT-6 Luna reaches \LuBrtBareAcc{} and \LuBrtBareFone{}, placing Luna level with JEV on macro F1 and behind on accuracy.  Qwen3.8-27B's accuracy (0.680) and macro F1 (0.518) are both lower than JEV's ($-0.034$ [$-0.037$, $-0.032$] and $-0.038$ [$-0.049$, $-0.026$], respectively) and lower than GPT-6 Luna's. Qwen marks 15 BBC articles as conflict and finds 6 of the 53 conflict articles, for a conflict F1 of 0.18, compared with 0.32 for JEV and 0.14 for Luna. ConfliBERT remains the stronger classifier, but it was fine-tuned on the labelled GTD incidents up to 2016, a training set that researchers starting a new task will rarely have. 

\subsection*{Open-ended survey responses}
\label{sec:survey}

\citet{digiuseppe2026} show that scales built from LLM pairwise comparisons agree more closely across models than naive (0--10) point-wise ratings do. Their first study scales 1,402 explanations of how interest rates are determined as a proxy for economic knowledge. Their second study scales uncertainty in respondent-supplied likely consequences of a potential breach of the US debt ceiling in 2023.\footnote{Both studies use nonprobability samples of US adults recruited on Prolific, with quotas on age, gender and partisanship in Study 1 and on age, sex and race or ethnicity in Study 2. Study 1 reuses the open-ended answers collected by \citet{digiuseppe2025cbi} in the first wave of their survey, fielded on 6 August 2024, keeping the 1,402 respondents who said they had not looked up either answer, and compares them in 28,040 published pairs, twenty per respondent. Study 2 is a survey experiment fielded on 16 May 2023 that randomly assigned respondents to a control, a certain or an uncertain arm before asking what would happen if the debt ceiling were not raised (1,514 answers in the archived files, 502, 505 and 507 by arm; the published articles report 1,486) \citep{digiuseppe2025information}. The human benchmark comes from respondents whose answers had been rated most knowledgeable and who were recontacted to compare pairs of answers. The archived files hold 460 judgments of 243 distinct pairs by 23 raters, each judging 20 pairs drawn from a pool of 300; the article reports 15 raters after screening and 300 rated pairs.} Both fit a Bayesian Bradley-Terry model \citep{bradley1952} to recover the scale from twenty comparisons per answer. We compare JEV with GPT-4o, the frontier model at the time of their study, and with the two contemporary models. 

Each study compares several language models, of varying sizes, on three tasks. The models judge the answers in pairs (there are 28,040 pairs in Study 1 and 30,280 in Study 2, with each answer paired with twenty others selected at random), rate each answer on a scale of 0 to 10 and, in Study 1 only, judge the pairs that the experts also judged. For each of these tasks we use the same pairs of answers and the same wording as the original, except that for JEV we leave out the lines about how the reply should be formatted. JEV judges each pair in both orders, with the published order as the primary specification. GPT-6 Luna and Qwen judge pairs only in the published order, except for the pairs judged by the experts, which they judge in both orders. For the ratings, we ask JEV to give each answer a Score on a ten-level scale and rescale these to the range of the original ratings. We also add two blocks of requests that are not in the original. The first has every pair among a random selection of 80 answers per study, in order to test transitivity. In the second we repeat requests, in order to test reliability.

The two contemporary models come as close to GPT-4o's scale as JEV does. In Study 1, the scales they produce correlate with the GPT-4o scale at 0.935 for GPT-6 Luna and 0.942 for Qwen3.8-27B. JEV's corresponding correlation is level with both (JEV minus Luna 0.002 [$-0.004$, 0.008] and JEV minus Qwen $-0.004$ [$-0.009$, 0.000]). Agreement among models can also reflect shared errors, so we turn to human judges. On the \HumanNPairs{} expert-judged pairs (Table~\ref{tab:f1}), GPT-4o reaches F1 = \GptFoFone{} and JEV \JevFonePrimary{} [\JevFoneCILo{}, \JevFoneCIHi{}]. The paired gap (GPT-4o minus JEV) of \GptFoFoneGap{} [\GptFoFoneGapLo{}, \GptFoFoneGapHi{}] includes zero.  GPT-6 Luna has an F1 of 0.817 [0.772, 0.856] and Qwen3.8-27B 0.782 [0.721, 0.831]. Both are level with JEV. In Study 2, compared with the certain arm, respondents in the uncertain arm score \JevSTwo{} [\JevSTwoLo{}, \JevSTwoHi{}] standard deviations higher on JEV's scale and \GptFoSTwo{} [\GptFoSTwoLo{}, \GptFoSTwoHi{}] on GPT-4o's.  The contemporary models agree with this result, giving 0.239 [0.109, 0.361] on the GPT-6 Luna scale and 0.239 [0.110, 0.362] on the Qwen3.8-27B scale.

\subsection*{Manifestos and speeches}
\label{sec:lemens}

\citet{lemens2025} use LLMs to position political texts on policy and ideological dimensions. They prompt an LLM to score each tweet or sentence on a specified scale, allowing an ``NA'' response when the text lacks relevant political content. For longer documents, they average the numeric sentence scores to estimate the document's position. We repeat this with JEV, GPT-6 Luna and Qwen3.8-27B for \LmgUkSent{} sentences from \LmgUkDocs{} British party manifestos on economic and social dimensions, and \LmgEuSent{} sentences from \LmgEuDocs{} European Parliament speeches on coal subsidies. The benchmarks are document-level judgments. The manifesto targets are expert coding and the speech targets are crowd coding from \citet{benoit2016}.

Both corpora are taken from \citet{benoit2016}. The manifesto corpus contains the manifestos of the Conservatives, Labour and the Liberal Democrats at the six general elections from 1987 to 2010, 18,263 sentences in total. Each sentence was coded by four to six political scientists (faculty members and doctoral students), which gives about seven ratings per sentence once both coding orders are counted. The benchmark is the scaled expert position of each manifesto, as calculated by \citet{benoit2016}. The speech corpus contains the 36 contributions to the European Parliament's debate of 23 November 2010 on state aid to close uncompetitive coal mines, 429 sentences in their original languages. Crowd workers coded these in six language versions, and the benchmark is the average of the six crowd scores.

\citet{lemens2025} report four applications. These are tweets from members of the 118th US Congress, the positions of the senators of the 117th Congress based on their tweets, the British manifestos and the coal debate. We replicate the last two, using the same specifications as their main results. The manifestos are used without a description of the dimension, and the speeches in their original languages with a description. For JEV, instead of the 0--100 scale we use nine worded Score levels, and instead of the instruction to reply NA we add a separate yes/no question asking whether the sentence is relevant. We fixed the primary inclusion threshold at 0.5 before the run. For GPT-6 Luna and Qwen we use the published prompts, including the scale and the NA option.

JEV's expected scores, averaged over included sentences, correlate \LmgEconJev{} with the economic benchmark and \LmgSocJev{} with the social benchmark, within the published models' range but below the published GPT-4o results of \LmgEconGpt{} and \LmgSocGpt{}, respectively. The speeches are harder. JEV reaches \LmgEuJev{} on the \LmgEuN{} speeches with at least one sentence it judges relevant, below GPT-4o's \LmgEuGpt{} (Table~\ref{tab:lemens}).  GPT-6 Luna is above JEV on all three benchmarks (correlations of 0.977, 0.952 and 0.947), while Qwen3.8-27B is level with JEV on all three (0.974, 0.940 and 0.905, judged level by a paired bootstrap over documents).

\subsection*{Party positions from party names}
\label{sec:dileo_summary}

\citet{dileo2025} test whether an LLM can place European parties on a left--right scale from its own embedded knowledge. This allows for scaling of parties that have no common sources of votes or corpora. In the application, the model sees only two party names and must decide which is more right-wing. A Bradley-Terry model turns these pairwise answers into a scale, which is compared with Chapel Hill expert placements  \citep{jolly2022}. 

The party names and benchmarks are from three sources. These are the Manifesto Project dataset \citep{lehmann2024}, the Chapel Hill Expert Survey \citep{jolly2022}, extended back to 1984 with the Ray--Marks--Steenbergen file \citep{ray1999,steenbergen2007}, whose left--right values come from earlier expert surveys for 1984 and 1996 and are interpolated for 1988 and 1992, and voters' placements of parties in the harmonised True European Voter surveys \citep{schmitt2021tev}. Placements from the Comparative Study of Electoral Systems \citep{cses2024} provide an alternative benchmark. We have one sample per European Parliament election between 1979 and 2019. Each sample includes the European parties that ran in a national election in the four years before that European Parliament election. The samples contain from 112 to 362 parties from 18 to 39 countries. Each party is compared with every other party in its sample, which gives 347,801 dyads in total. From 1984 on there are such placements for between 74 and 239 parties a year (those used for 1989 are mostly interpolations), and for 1979 there are none.

The task covers the whole of the main analysis in the original, which concerns the left--right dimension only, with every dyad in every year. \citet{dileo2025} take the most common answer of seven runs. We use only one run. For JEV we remove the closing instruction to reply with exactly ``Party 1'' or ``Party 2'', but we keep the original order of presentation. We do not repeat the robustness analyses in their supplementary material, such as pooled years, few-shot prompts or other framings.

Across the \LuChesN{} election years, JEV's scale correlates on average (Table~\ref{tab:dileo}) \DilChesMean{} with the experts, above the \DilChesTheirMean{} of the GPT-3.5 model in the original study, although their supplement reports that GPT-4o and Llama-3.1 70B did better than GPT-3.5. GPT-6 Luna averages \LuChesMean{}, and its difference from JEV is indistinguishable from zero in all \LuChesN{} years.  Qwen3.8-27B averages 0.755 and is below JEV in all eight years, with differences of between 0.09 and 0.15. 

\subsection*{Democracy indicators from country names}
\label{sec:vdem_summary}

\citet{weidmann2026} ask whether LLMs can reproduce V-Dem expert codes. The model receives a country's name, the year and a V-Dem question with its answer scale, and returns a code. We repeat this for \VdIndicators{} indicators in \VdCountries{} countries in 2023. JEV's codes correlate \VdJevCor{} with V-Dem's on average (Table~\ref{tab:vdem}), above GPT-4o (\VdGptCor{}) and Llama-3.1 70B (\VdLlamaCor{}), recomputed from the original study's published codes (the article reports 0.64 for GPT-4o), and above GPT-6 Luna (\VdLunaCor{}) and Qwen3.8-27B (0.59). All models do worse on the codes that changed from the previous year.

We use version 14 of V-Dem \citep{coppedge2024dataset,coppedge2024}, coded for 2023, as the benchmark. The 53 indicators are the ordinal, expert-coded indicators that \citet{weidmann2026} selected from the components of V-Dem's five high-level democracy indices; indicators coded only in election years are not among them. Each code is based on the ratings of several country experts (a median of six per code in V-Dem's coder-level data, \citealt{coppedge2024b}) and is computed with V-Dem's measurement model, which adjusts for differences between experts in how they use the scale \citep{pemstein2024}. Of the 9,063 country-indicator pairs, 9,041 have a V-Dem code and are scored.

In a single task, \citet{weidmann2026} code all 53 indicators with GPT-4o and Llama-3.1 70B, and then analyse how the models deviate systematically from V-Dem by country and as a function of disagreement between coders. We run the full task. For JEV, we use the V-Dem question as the instruction and ``What is the score for [country] in 2023?'' as the input. The ordered response scale gives the levels of a Score, and we remove the instruction to reply with a number. For GPT-6 Luna and Qwen we use the published prompt.

In both recall tasks the models answer from what they learned in training, not from a text they read. The expert placements and V-Dem codes were public before these models were trained, so high agreement may partly reflect recall of the benchmarks themselves. This matters most for V-Dem: \citet{weidmann2026} used version 14 because it was released after GPT-4o and Llama-3.1 were trained, but JEV, GPT-6 Luna and Qwen3.8-27B were trained after its release, so the comparison with the published models may favour the newer models.

In the V-Dem task, JEV has a small but clear lead over every model. Its mean correlation with V-Dem is higher than GPT-6 Luna's by 0.05 [0.04, 0.07], GPT-4o's by 0.05 [0.03, 0.06], Qwen3.8-27B's by 0.09 [0.07, 0.11] and Llama-3.1 70B's by 0.18 [0.16, 0.20].

\section*{JEV's advantage?}
\label{sec:weighting}

JEV annotates and scales text in these political science contexts about as well as a mid-tier commercial LLM and, in many instances, as well as open-weight models that can be run locally. Further, JEV does not show much of a cost advantage over what is commercially available in autumn 2026. JEV does have a clear speed advantage. However, for most scholarly uses, speed is not a deciding factor. On these metrics, there is little reason to adopt JEV in place of traditional, token-based large language models. 

Beyond these metrics, JEV's supposed calibration advantage might be another reason to prefer it over traditional LLMs. Probabilities help scholars in two ways. First, they rank texts by how likely each answer is, so researchers can screen a corpus for a rare category, or send the answers the model is least sure of to a human or a more capable (and expensive) LLM. For this the probabilities need to rank answers well, which AUROC measures. Second, researchers can weight each answer by its probability to build a continuous scale \citep{licht2025}, or read the probabilities as rates. For this they also need to be calibrated, which ECE measures. Both uses rely on the probabilities being informative and available. 

\begin{figure}[htbp]
\centering
\includegraphics[width=\textwidth]{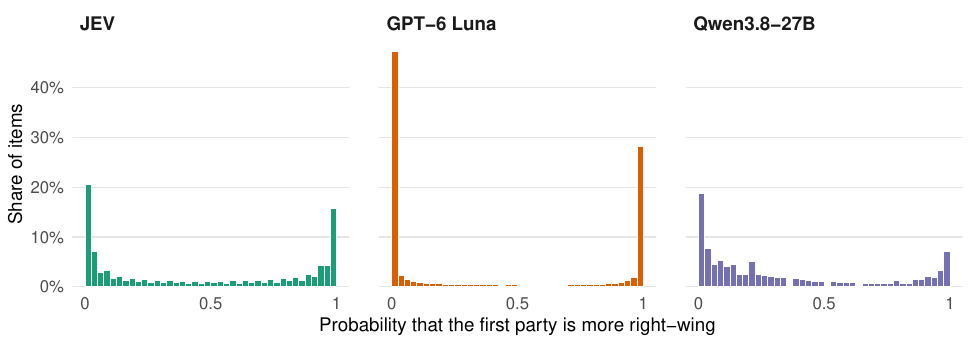}
\caption{Distribution of each model's probability that the first party is the more right-wing, across all party comparisons in our replication of \citet{dileo2025} ($N$ = 347,801; one Qwen reply was lost). For JEV the probability comes from a two-option Choice; for GPT-6 Luna and Qwen3.8-27B, from their first-token probabilities.}
\label{fig:dist_fig1}
\end{figure}

As we noted in the introduction, some providers now report few token probabilities or none, and those GPT-6 Luna reports are mostly close to 0 or 1. Figure~\ref{fig:dist_fig1} shows this for the party comparisons in our replication of \citet{dileo2025}: 74\% of Luna's probabilities are below 0.01 or above 0.99, against 18\% for JEV and 16\% for Qwen. JEV returns more mid-range probabilities, as we would expect when two parties are close ideologically. Qwen does the same, so near-certainty is not a property of LLMs as such. 

\begin{figure}[htbp]
\centering
\includegraphics[width=\textwidth]{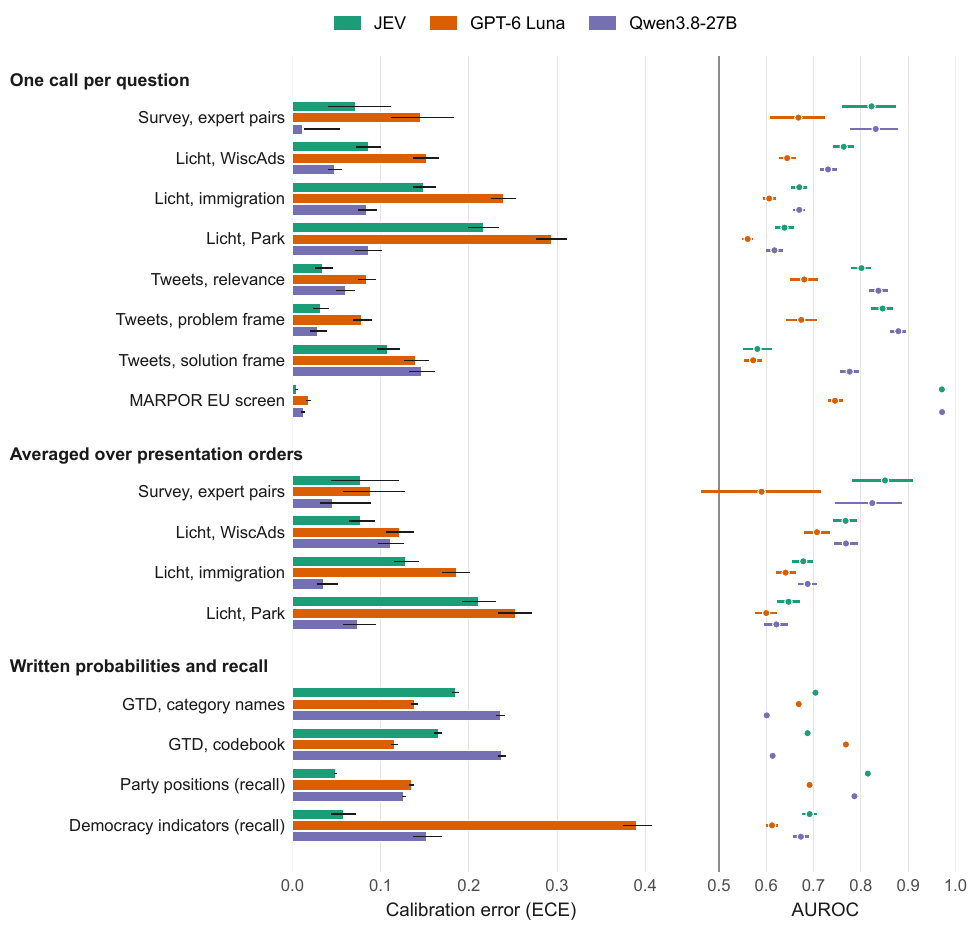}
\caption{Calibration of JEV, GPT-6 Luna and Qwen3.8-27B against human coders, by task, with 95\% bootstrap intervals. Left: calibration error (ECE), the gap between the probability a model gives its answer and the share of such answers that agree with the human label, over ten probability bins; lower is better. Right: AUROC, the chance that a correct answer gets a higher probability than an incorrect one; 0.5 is chance. Top rows: each question asked once. Middle rows: probabilities averaged over the two presentation orders. Bottom rows: on attack type GPT-6 Luna and Qwen write their probabilities in the reply; party positions and democracy indicators are recall tasks. In the attack-type and party-position rows, which have tens or hundreds of thousands of items, the intervals are narrower than the markers. Appendix H describes the Licht and MARPOR tasks.}
\label{fig:calib_luna}
\end{figure}

The difference in probabilities is indicative, but not conclusive, evidence of poor calibration in GPT-6 Luna. Figure~\ref{fig:calib_luna} presents the expected calibration error (ECE, left panel; \citealt{naeini2015,guo2017}) and the area under the receiver operating characteristic curve (AUROC, right panel; \citealt{hanley1982}). The former groups each model's answers by its stated probability, compares that probability with the share of answers that agree with the human coders in each group, and averages the gaps, weighting by group size. An ECE of zero, for example, means that answers given with 80\% probability agree with the human coders 80\% of the time. An ECE of one would mean the model is 100\% confident and got every answer wrong.

When each question is asked once, GPT-6 Luna has higher calibration error than JEV on all eight reading tasks with a categorical human label. This suggests that TypeSafe's claims of better calibration might be warranted against Luna. The gap disappears on the open-ended survey pairs when we average over the two presentation orders, and it reverses on the attack-type task, where Luna writes its probabilities in its reply. Qwen3.8-27B's calibration differs from JEV's by type of task. Its calibration error is lower on the four pairwise tasks, higher on three of the four labelling tasks, and higher on both attack-type arms and both recall tasks. JEV's advantage therefore holds relative to Luna's token probabilities but not consistently relative to Qwen's. Yet, Where JEV does show better calibration than Qwen, the differences are not dramatic. 

The AUROC presents a similar picture. It is the chance that a randomly chosen right answer gets a higher probability than a randomly chosen wrong one. Luna's AUROC is lower than JEV's on seven of the eight tasks and level on the other, and lower than Qwen's on all eight. On attack type, where it writes its probabilities, Luna's AUROC is above Qwen's in both arms and above JEV's only when given the codebook. JEV and Qwen are not consistently different. 

\section*{Conclusion}

Our analysis indicates that JEV is a capable model for text scaling and annotation in political science contexts. However, there are no major gains in cost or capabilities that make it a clear choice for most applications. The main advantages lie in its speed and ease of parsing the underlying probabilities. When each question is asked once, JEV's probabilities are also better calibrated than GPT-6 Luna's token probabilities, though not consistently better than those of Qwen3.8-27B. As such, the choice to use JEV over other LLMs is task dependent. In some cases, JEV might offer more accurate calibration while still performing on par with LLMs for social science tasks. Our analysis suggest that this still requires that researchers validate JEV as they would other LLMs.

\clearpage

\bibliographystyle{chicago}
\bibliography{refs}

\clearpage

\section*{Use of AI tools}

We used Claude Opus 5 and Opus 5.5 (Anthropic) to generate analysis code and Fable 5.1 (Anthropic) to check the code against reported figures. GPT-6 Astra (OpenAI) drafted the initial manuscript, which the authors heavily edited and added to. Before posting, we used Claude Opus 5.5 to audit the paper's claims and citations against their sources and our results, and for copy editing. Claude Sonnet 5.5 proofread the text. We reviewed every resulting change. All design choices, analysis and conclusions are ours, and we take full responsibility for the content.

\appendix
\section*{Appendix A: prompts as sent}

The survey application's prompts are given first, followed by those of the
other applications, each copied from the first request of its kind in the
files that were sent.

\subsection*{Study 1: interest-rate knowledge}

\textbf{Open-ended item.} In a few sentences and without looking it up, can you explain how interest rates (i.e. the cost of borrowing money to buy a house or car) go up or down in the US economy?

\paragraph{Pairwise Choice with a tie option (Choice).}

\emph{Instructions.} You are an expert in US economic policy. Your task is to determine which of two given statements contains a more knowledgeable response to the following question: In a few sentences and without looking it up, can you explain how interest rates (i.e. the cost of borrowing money to buy a house or car) go up or down in the US economy?

\emph{Options.}
\begin{itemize}
\item \texttt{first}: The first statement contains more knowledge.
\item \texttt{second}: The second statement contains more knowledge.
\item \texttt{equal}: They are equal or incomparable.
\end{itemize}

\paragraph{Pairwise Choice without a tie option (Choice).}

\emph{Instructions.} You are an expert in US economic policy. Your task is to determine which of two given statements contains a more knowledgeable response to the following question: In a few sentences and without looking it up, can you explain how interest rates (i.e. the cost of borrowing money to buy a house or car) go up or down in the US economy? If both are equally knowledgeable choose one with a slight preference.

\emph{Options.}
\begin{itemize}
\item \texttt{first}: The first statement contains more knowledge.
\item \texttt{second}: The second statement contains more knowledge.
\end{itemize}

\paragraph{Pairwise Noul (Noul).}

\emph{Instructions.} You are an expert in US economic policy. Does the first statement contain a more knowledgeable response than the second statement, in response to the following question: In a few sentences and without looking it up, can you explain how interest rates (i.e. the cost of borrowing money to buy a house or car) go up or down in the US economy?

\emph{Options.}
\begin{itemize}
\item \texttt{true}: The first statement contains more knowledge.
\item \texttt{false}: The second statement contains more knowledge.
\end{itemize}

\paragraph{Score, ten levels (Score).}

\emph{Instructions.} You are an expert in US economic policy. Your task is to rate the given statement based on how knowledgeable it is in response to the following question: In a few sentences and without looking it up, can you explain how interest rates (i.e. the cost of borrowing money to buy a house or car) go up or down in the US economy?

\emph{Levels, low to high.}
\begin{enumerate}\setcounter{enumi}{-1}
\item Completely incorrect or irrelevant.
\item Almost entirely wrong, with at most a stray relevant word.
\item Mostly wrong, with a trace of relevant knowledge.
\item Largely uninformed, but gestures at something relevant.
\item Partly correct, with significant errors or omissions.
\item Roughly correct in outline, but vague or incomplete.
\item Correct on the main point, with minor gaps.
\item Accurate and reasonably complete.
\item Accurate, complete, and clearly explained.
\item Highly knowledgeable and accurate.
\end{enumerate}

\paragraph{Score, five levels (Score).}

\emph{Instructions.} You are an expert in US economic policy. Your task is to rate the given statement based on how knowledgeable it is in response to the following question: In a few sentences and without looking it up, can you explain how interest rates (i.e. the cost of borrowing money to buy a house or car) go up or down in the US economy?

\emph{Levels, low to high.}
\begin{enumerate}\setcounter{enumi}{-1}
\item Completely incorrect or irrelevant.
\item Mostly wrong, with a trace of relevant knowledge.
\item Partly correct, with significant errors or omissions.
\item Correct on the main point, with minor gaps.
\item Highly knowledgeable and accurate.
\end{enumerate}

\paragraph{Noul for a single answer (Noul).}

\emph{Instructions.} You are an expert in US economic policy. Is the statement a knowledgeable and accurate response to the following question: In a few sentences and without looking it up, can you explain how interest rates (i.e. the cost of borrowing money to buy a house or car) go up or down in the US economy?

\emph{Options.}
\begin{itemize}
\item \texttt{true}: Highly knowledgeable and accurate.
\item \texttt{false}: Completely incorrect or irrelevant.
\end{itemize}

\subsection*{Study 2: debt-ceiling uncertainty}

\textbf{Open-ended item.} In one or two sentences, what do you think will happen if the government DOES NOT increase the debt ceiling?

\paragraph{Pairwise Choice with a tie option (Choice).}

\emph{Instructions.} Right before the 2023 Debt Ceiling Crisis in the United States we asked US citizens the following question: In one or two sentences, what do you think will happen if the government DOES NOT increase the debt ceiling? Your job is to compare each statement and determine which one exhibits the greatest uncertainty about what might happen

\emph{Options.}
\begin{itemize}
\item \texttt{first}: The first statement contains more uncertain.
\item \texttt{second}: The second statement is more uncertain.
\item \texttt{equal}: They have the same level of certianty or incomparable.
\end{itemize}

\paragraph{Pairwise Noul (Noul).}

\emph{Instructions.} Right before the 2023 Debt Ceiling Crisis in the United States we asked US citizens the following question: In one or two sentences, what do you think will happen if the government DOES NOT increase the debt ceiling? Does the first statement exhibit greater uncertainty about what might happen than the second statement?

\emph{Options.}
\begin{itemize}
\item \texttt{true}: The first statement is more uncertain.
\item \texttt{false}: The second statement is more uncertain.
\end{itemize}

\paragraph{Score, ten levels (Score).}

\emph{Instructions.} Right before the 2023 Debt Ceiling Crisis in the United States we asked US citizens the following question: In one or two sentences, what do you think will happen if the government DOES NOT increase the debt ceiling? Your job is to read one of the answers and rank the statement based on the degree of uncertainty about what might happen

\emph{Levels, low to high.}
\begin{enumerate}\setcounter{enumi}{-1}
\item Completely uncertain about what might happen.
\item Almost entirely uncertain, naming nothing concrete.
\item Mostly uncertain, with one vague possibility raised.
\item Largely hedged, though some direction is suggested.
\item Mixed, with uncertainty outweighing any stated expectation.
\item Mixed, with a stated expectation outweighing the hedging.
\item Fairly certain, with some hedging remaining.
\item Certain about the main consequence, with minor hedging.
\item Confident and specific about what will happen.
\item Highly certain about what will happen.
\end{enumerate}

\paragraph{Score, five levels (Score).}

\emph{Instructions.} Right before the 2023 Debt Ceiling Crisis in the United States we asked US citizens the following question: In one or two sentences, what do you think will happen if the government DOES NOT increase the debt ceiling? Your job is to read one of the answers and rank the statement based on the degree of uncertainty about what might happen

\emph{Levels, low to high.}
\begin{enumerate}\setcounter{enumi}{-1}
\item Completely uncertain about what might happen.
\item Mostly uncertain, with one vague possibility raised.
\item Mixed between uncertainty and a stated expectation.
\item Fairly certain, with some hedging remaining.
\item Highly certain about what will happen.
\end{enumerate}

\paragraph{Noul for a single answer (Noul).}

\emph{Instructions.} Right before the 2023 Debt Ceiling Crisis in the United States we asked US citizens the following question: In one or two sentences, what do you think will happen if the government DOES NOT increase the debt ceiling? Is the statement certain about what will happen?

\emph{Options.}
\begin{itemize}
\item \texttt{true}: Highly certain about what will happen.
\item \texttt{false}: Completely uncertain about what might happen.
\end{itemize}

\subsection*{Gilardi et al.: tweet relevance and framing}

The state is the tweet text. The three questions were bundled in one request.

\paragraph{Relevance (Choice).}

\emph{Instructions.} In this job, you will be shown a sample of Tweets collected from the social media platform Twitter. Your task will be to determine if the Tweets have to do with “content” moderation” or not. \newline “Content moderation” refers to the practice of screening and monitoring content posted by users on social media sites to determine if the content should be published or not, based on specific rules and guidelines. \newline Every time someone posts something on a platform like Facebook or Twitter, that piece of content goes through a review process (‘content moderation’) to ensure that it is not illegal, hateful or inappropriate and that it complies with the rules of the site. When that is not the case, that piece of content can be removed, flagged, labelled as or ‘disputed’. \newline Deciding what should be allowed on social media is not always easy. For example, many sites ban child pornography and terrorist content as it is illegal. However, things are less clear when it comes to content about the safety of vaccines or politics, for example. Even when people agree that some content should be blocked, they do not always agree about the best way to do so, about how effective it is and who should do it (the government or private companies, human moderators or artificial intelligence). \newline For each tweet in the sample: Carefully read the text of the Tweet, paying close attention to details. Classify the Tweet as either irrelevant (0) or relevant (1). \newline Tweets should be coded as relevant when they directly relate to content moderation. This includes Tweets that discuss social media platforms’ content moderation rules and practices, and Tweets that discuss governments’ regulation of online content moderation. This also includes Tweets that discuss mild forms of content moderation, like flagging Tweets and Tweets when they indirectly relate to content moderation. \newline Tweets should be coded as irrelevant if they do not refer to content moderation or if they are themselves examples of moderated content. This would include, for example, a Tweet by Donald Trump that Twitter has labelled as ‘disputed’, a Tweet claiming that something is false, or a Tweet containing sensitive content.

\emph{Options.}
\begin{itemize}
\item \texttt{relevant}: The Tweet directly relates to content moderation, including platforms' content moderation rules and practices, governments' regulation of online content moderation, and mild forms such as flagging.
\item \texttt{irrelevant}: The Tweet does not refer to content moderation, or is itself an example of moderated content.
\end{itemize}

\paragraph{Problem frame (Choice).}

\emph{Instructions.} In this job, you will be shown a sample of Tweets collected from the social media platform Twitter. “Content moderation” refers to the practice of screening and monitoring content posted by users on social media sites to determine if the content should be published or not, based on specific rules and guidelines. Your task is to determine whether the Tweet frames content moderation as a PROBLEM. Tweets should be coded as framing content moderation as a problem when they emphasise negative effects of content moderation, for instance on free speech, or when they describe content moderation as biased, censorious, ineffective or excessive.

\emph{Options.}
\begin{itemize}
\item \texttt{yes}: The Tweet frames content moderation as a problem.
\item \texttt{no}: The Tweet does not frame content moderation as a problem.
\end{itemize}

\paragraph{Solution frame (Choice).}

\emph{Instructions.} In this job, you will be shown a sample of Tweets collected from the social media platform Twitter. “Content moderation” refers to the practice of screening and monitoring content posted by users on social media sites to determine if the content should be published or not, based on specific rules and guidelines. Your task is to determine whether the Tweet frames content moderation as a SOLUTION. Tweets should be coded as framing content moderation as a solution when they emphasise positive effects of content moderation, for instance in reducing harmful, hateful or misleading content, or when they call for more or better moderation.

\emph{Options.}
\begin{itemize}
\item \texttt{yes}: The Tweet frames content moderation as a solution.
\item \texttt{no}: The Tweet does not frame content moderation as a solution.
\end{itemize}

\subsection*{Ornstein et al.: tweet sentiment}

The state is the tweet text. The few-shot form appends their six examples per case to the zero-shot instruction; both are shown for each case.

\paragraph{Sentiment, zero-shot (Choice).}

\emph{Masterpiece Cakeshop}

\emph{Instructions.} Read these tweets posted the day after the US Supreme Court ruled in favor of a baker who refused to bake a wedding cake for a same-sex couple. For each tweet, decide whether its sentiment is Positive, Neutral, or Negative.

\emph{Options.}
\begin{itemize}
\item \texttt{Positive}: The tweet expresses positive sentiment about the Court's decision.
\item \texttt{Neutral}: The tweet expresses neither positive nor negative sentiment about the Court's decision.
\item \texttt{Negative}: The tweet expresses negative sentiment about the Court's decision.
\end{itemize}

\paragraph{Sentiment, few-shot (Choice).}

\emph{Masterpiece Cakeshop}

\emph{Instructions.} Read these tweets posted the day after the US Supreme Court ruled in favor of a baker who refused to bake a wedding cake for a same-sex couple. For each tweet, decide whether its sentiment is Positive, Neutral, or Negative.

Here are six labelled examples.

Tweet: Thank you Supreme Court I take pride in your decision!!!![emoji] \#SCOTUS \newline Sentiment: Positive

Tweet: Supreme Court rules in favor of Colorado baker! This day is getting better by the minute! \newline Sentiment: Positive

Tweet: Can’t escape the awful irony of someone allowed to use religion to discriminate against people in love. \newline Not my Jesus. \newline \#opentoall \#SCOTUS \#Hypocrisy \#MasterpieceCakeshop \newline Sentiment: Negative

Tweet: I can’t believe this cake case went all the way to \#SCOTUS . Can someone let me know what cake was ultimately served at the wedding? Are they married and living happily ever after? \newline Sentiment: Neutral

Tweet: Supreme Court rules in favor of baker who would not make wedding cake for gay couple \newline Sentiment: Neutral

Tweet: \#SCOTUS set a dangerous precedent today. Although the Court limited the scope to which a business owner could deny services to patrons, the legal argument has been legitimized that one's subjective religious convictions trump (no pun intended) \#humanrights. \#LGBTQRights \newline Sentiment: Negative

\emph{Options.}
\begin{itemize}
\item \texttt{Positive}: The tweet expresses positive sentiment about the Court's decision.
\item \texttt{Neutral}: The tweet expresses neither positive nor negative sentiment about the Court's decision.
\item \texttt{Negative}: The tweet expresses negative sentiment about the Court's decision.
\end{itemize}

\paragraph{Sentiment, zero-shot (Choice).}

\emph{Mazars}

\emph{Instructions.} Read these tweets posted the day after the US Supreme Court ruled that sitting presidents are not immune to state criminal subpoenas, and that President Trump was obliged to disclose his tax returns to the Manhattan District Attorney. For each tweet, decide whether its sentiment is Positive, Neutral, or Negative.

\emph{Options.}
\begin{itemize}
\item \texttt{Positive}: The tweet expresses positive sentiment about the Court's decision.
\item \texttt{Neutral}: The tweet expresses neither positive nor negative sentiment about the Court's decision.
\item \texttt{Negative}: The tweet expresses negative sentiment about the Court's decision.
\end{itemize}

\paragraph{Sentiment, few-shot (Choice).}

\emph{Mazars}

\emph{Instructions.} Read these tweets posted the day after the US Supreme Court ruled that sitting presidents are not immune to state criminal subpoenas, and that President Trump was obliged to disclose his tax returns to the Manhattan District Attorney. For each tweet, decide whether its sentiment is Positive, Neutral, or Negative.

Here are six labelled examples.

Tweet: Justice \#ClarenceThomas is waste of space on the \#scotus \newline Sentiment: Negative

Tweet: BREAKING: Supreme Court Justice Ruth Bader Ginsburg has been hospitalized for a possible infection, per a SCOTUS spokesperson. @Scotus @ruthbadergins \newline Sentiment: Neutral

Tweet: The Supreme Court is going to disappoint us tomorrow. And trump will feel even more untouchable. He’ll brag about it at his Klan rallies. Sweaty orange spray tan pooling above his lip, smug faced as he gloats and brags. It makes me sick. \newline Sentiment: Negative

Tweet: SCOTUS just ruled Manhattan DA CAN get trumps financials and tax returns.  This is a great day for the ruke of law and America. \newline Sentiment: Positive

Tweet: Today the Supreme Court let @realDonaldTrump know that he is not above the law! \newline Sentiment: Positive

Tweet: Both SCOTUS rulings in Trump financial records sent back to lower courts. Practically speaking that means no turnover of records immediately in either case. \#7News \newline Sentiment: Neutral

\emph{Options.}
\begin{itemize}
\item \texttt{Positive}: The tweet expresses positive sentiment about the Court's decision.
\item \texttt{Neutral}: The tweet expresses neither positive nor negative sentiment about the Court's decision.
\item \texttt{Negative}: The tweet expresses negative sentiment about the Court's decision.
\end{itemize}

\subsection*{Di Leo et al.: party positions}

The state lists the two parties by name in their original language with their countries, in the order of the original duplet. The year in the instruction is the reference year; the 1979 request is shown. Their system prompt is reproduced verbatim.

\emph{State, first request.} Party 1: Freiheitliche Partei Österreichs Party 1 country: Austria Party 2: Sozialdemokratische Partei Österreichs Party 2 country: Austria

\paragraph{More right-wing (Choice).}

\emph{Instructions.} In political matters, people talk of 'the Left' and 'the Right'. You will be given the names of two parties in their original language, and the country of origin of each party. Based on their overall ideological stance, which one of these two parties was more right-wing in year 1979?

\emph{Options.}
\begin{itemize}
\item \texttt{Party 1}: Party 1 was more right-wing.
\item \texttt{Party 2}: Party 2 was more right-wing.
\end{itemize}

\subsection*{Brandt et al.: attack type}

The state is the incident text as their code builds it (summary, then motive). Only the first incident's text is shown; every design used the same text.

\emph{State, first incident.} 01/02/2017: Assailants abducted eleven construction laborers building a school in Takhta Pul, Kandahar, Afghanistan. The hostages were released on January 20, 2017. No group claimed responsibility for the incident; however, sources attributed the kidnapping to the Taliban.

\paragraph{Nine-way Choice, category names only (Choice).}

\emph{Instructions.} Classify the following event into these categories: \newline Assassination, Armed Assault, Bombing/Explosion, Hijacking, \newline Hostage Taking (Barricade Incident), Hostage Taking (Kidnapping), \newline Facility/Infrastructure Attack, Unarmed Assault, Unknown

\emph{Options.}
\begin{itemize}
\item \texttt{Assassination}: The attack type is Assassination.
\item \texttt{Armed Assault}: The attack type is Armed Assault.
\item \texttt{Bombing/Explosion}: The attack type is Bombing/Explosion.
\item \texttt{Hijacking}: The attack type is Hijacking.
\item \texttt{Hostage Taking (Barricade Incident)}: The attack type is Hostage Taking (Barricade Incident).
\item \texttt{Hostage Taking (Kidnapping)}: The attack type is Hostage Taking (Kidnapping).
\item \texttt{Facility/Infrastructure Attack}: The attack type is Facility/Infrastructure Attack.
\item \texttt{Unarmed Assault}: The attack type is Unarmed Assault.
\item \texttt{Unknown}: The attack type is Unknown.
\end{itemize}

\paragraph{Nine-way Choice with the codebook (Choice).}

\emph{The option descriptions are the GTD codebook's definitions, verbatim.}

\emph{Instructions.} Classify the following event into these categories: \newline Assassination, Armed Assault, Bombing/Explosion, Hijacking, \newline Hostage Taking (Barricade Incident), Hostage Taking (Kidnapping), \newline Facility/Infrastructure Attack, Unarmed Assault, Unknown

Use the following definitions from the Global Terrorism Database Codebook.

This field captures the general method of attack and often reflects the broad class of tactics used. It consists of nine categories, which are defined below. Up to three attack types can be recorded for each incident. Typically, only one attack type is recorded for each incident unless the attack is comprised of a sequence of events.

When multiple attack types may apply, the most appropriate value is determined based on the hierarchy below. For example, if an assassination is carried out through the use of an explosive, the Attack Type is coded as Assassination, not Bombing/Explosion. If an attack involves a sequence of events, then the first, the second, and the third attack types are coded in the order of the hierarchy below rather than the order in which they occurred.

Attack Type Hierarchy: \newline Assassination \newline Hijacking \newline Kidnapping \newline Barricade Incident \newline Bombing/Explosion \newline Armed Assault \newline Unarmed Assault \newline Facility/Infrastructure Attack \newline Unknown

\emph{Options.}
\begin{itemize}
\item \texttt{Assassination}: An act whose primary objective is to kill one or more specific, prominent individuals. Usually carried out on persons of some note, such as high-ranking military officers, government officials, celebrities, etc. Not to include attacks on non-specific members of a targeted group. The killing of a police officer would be an armed assault unless there is reason to believe the attackers singled out a particularly prominent officer for assassination.
\item \texttt{Armed Assault}: An attack whose primary objective is to cause physical harm or death directly to human beings by use of a firearm, incendiary, or sharp instrument (knife, etc.). Not to include attacks involving the use of fists, rocks, sticks, or other handheld (less-than-lethal) weapons. Also includes attacks involving certain classes of explosive devices in addition to firearms, incendiaries, or sharp instruments. The explosive device subcategories that are included in this classification are grenades, projectiles, and unknown or other explosive devices that are thrown.
\item \texttt{Bombing/Explosion}: An attack where the primary effects are caused by an energetically unstable material undergoing rapid decomposition and releasing a pressure wave that causes physical damage to the surrounding environment. Can include either high or low explosives (including a dirty bomb) but does not include a nuclear explosive device that releases energy from fission and/or fusion, or an incendiary device where decomposition takes place at a much slower rate. \newline If an attack involves certain classes of explosive devices along with firearms, incendiaries, or sharp objects, then the attack is coded as an armed assault only. The explosive device subcategories that are included in this classification are grenades, projectiles, and unknown or other explosive devices that are thrown in which the bombers are also using firearms or incendiary devices.
\item \texttt{Hijacking}: An act whose primary objective is to take control of a vehicle such as an aircraft, boat, bus, etc. for the purpose of diverting it to an unprogrammed destination, force the release of prisoners, or some other political objective. Obtaining payment of a ransom should not the sole purpose of a Hijacking, but can be one element of the incident so long as additional objectives have also been stated. Hijackings are distinct from Hostage Taking because the target is a vehicle, regardless of whether there are people/passengers in the vehicle.
\item \texttt{Hostage Taking (Barricade Incident)}: An act whose primary objective is to take control of hostages for the purpose of achieving a political objective through concessions or through disruption of normal operations. Such attacks are distinguished from kidnapping since the incident occurs and usually plays out at the target location with little or no intention to hold the hostages for an extended period in a separate clandestine location.
\item \texttt{Hostage Taking (Kidnapping)}: An act whose primary objective is to take control of hostages for the purpose of achieving a political objective through concessions or through disruption of normal operations. Kidnappings are distinguished from Barricade Incidents (above) in that they involve moving and holding the hostages in another location.
\item \texttt{Facility/Infrastructure Attack}: An act, excluding the use of an explosive, whose primary objective is to cause damage to a non-human target, such as a building, monument, train, pipeline, etc. Such attacks include arson and various forms of sabotage (e.g., sabotaging a train track is a facility/infrastructure attack, even if passengers are killed). Facility/infrastructure attacks can include acts which aim to harm an installation, yet also cause harm to people incidentally (e.g. an arson attack primarily aimed at damaging a building, but causes injuries or fatalities).
\item \texttt{Unarmed Assault}: An attack whose primary objective is to cause physical harm or death directly to human beings by any means other than explosive, firearm, incendiary, or sharp instrument (knife, etc.). Attacks involving chemical, biological or radiological weapons are considered unarmed assaults.
\item \texttt{Unknown}: The attack type cannot be determined from the available information.
\end{itemize}

\paragraph{One question per type (Noul).}

\emph{Nine requests per incident, one per type; the Assassination request is shown and the others substitute the type's name.}

\emph{Instructions.} Classify the following event. Does it belong to the category "Assassination"?

\emph{Options.}
\begin{itemize}
\item \texttt{true}: The event belongs to the category "Assassination".
\item \texttt{false}: The event does not belong to the category "Assassination".
\end{itemize}

\paragraph{Qwen3.5-9B, label names only (their prompt, verbatim).}

Sent to \texttt{qwen/qwen3.5-9b} as a single user message, with JSON output, seed 66502, at most 300 tokens, reasoning effort \texttt{none}, and the provider pinned to deepinfra at bf16 with no fallback.

\emph{Message.} Classify the following event into up to three of these categories, providing probabilities for each: \newline Assassination, Armed Assault, Bombing/Explosion, Hijacking, \newline Hostage Taking (Barricade Incident), Hostage Taking (Kidnapping), \newline Facility/Infrastructure Attack, Unarmed Assault, Unknown

For the event, return only a single JSON object with category names as keys and probabilities as values. \newline Example format: \{"Armed Assault": 0.7, "Bombing/Explosion": 0.2, "Unknown": 0.1\}

Event: \newline "<incident text>"

\paragraph{Qwen3.5-9B, codebook.}

Sent to \texttt{qwen/qwen3.5-9b} as a single user message, with JSON output, seed 66502, at most 300 tokens, reasoning effort \texttt{none}, and the provider pinned to deepinfra at bf16 with no fallback.

\emph{Message.} Classify the following event into up to three of these categories, providing probabilities for each: \newline Assassination, Armed Assault, Bombing/Explosion, Hijacking, \newline Hostage Taking (Barricade Incident), Hostage Taking (Kidnapping), \newline Facility/Infrastructure Attack, Unarmed Assault, Unknown

Use the following definitions from the Global Terrorism Database Codebook.

This field captures the general method of attack and often reflects the broad class of tactics used. It consists of nine categories, which are defined below. Up to three attack types can be recorded for each incident. Typically, only one attack type is recorded for each incident unless the attack is comprised of a sequence of events.

When multiple attack types may apply, the most appropriate value is determined based on the hierarchy below. For example, if an assassination is carried out through the use of an explosive, the Attack Type is coded as Assassination, not Bombing/Explosion. If an attack involves a sequence of events, then the first, the second, and the third attack types are coded in the order of the hierarchy below rather than the order in which they occurred.

Attack Type Hierarchy: \newline Assassination \newline Hijacking \newline Kidnapping \newline Barricade Incident \newline Bombing/Explosion \newline Armed Assault \newline Unarmed Assault \newline Facility/Infrastructure Attack \newline Unknown

Assassination: An act whose primary objective is to kill one or more specific, prominent individuals. Usually carried out on persons of some note, such as high-ranking military officers, government officials, celebrities, etc. Not to include attacks on non-specific members of a targeted group. The killing of a police officer would be an armed assault unless there is reason to believe the attackers singled out a particularly prominent officer for assassination. \newline Armed Assault: An attack whose primary objective is to cause physical harm or death directly to human beings by use of a firearm, incendiary, or sharp instrument (knife, etc.). Not to include attacks involving the use of fists, rocks, sticks, or other handheld (less-than-lethal) weapons. Also includes attacks involving certain classes of explosive devices in addition to firearms, incendiaries, or sharp instruments. The explosive device subcategories that are included in this classification are grenades, projectiles, and unknown or other explosive devices that are thrown. \newline Bombing/Explosion: An attack where the primary effects are caused by an energetically unstable material undergoing rapid decomposition and releasing a pressure wave that causes physical damage to the surrounding environment. Can include either high or low explosives (including a dirty bomb) but does not include a nuclear explosive device that releases energy from fission and/or fusion, or an incendiary device where decomposition takes place at a much slower rate. \newline If an attack involves certain classes of explosive devices along with firearms, incendiaries, or sharp objects, then the attack is coded as an armed assault only. The explosive device subcategories that are included in this classification are grenades, projectiles, and unknown or other explosive devices that are thrown in which the bombers are also using firearms or incendiary devices. \newline Hijacking: An act whose primary objective is to take control of a vehicle such as an aircraft, boat, bus, etc. for the purpose of diverting it to an unprogrammed destination, force the release of prisoners, or some other political objective. Obtaining payment of a ransom should not the sole purpose of a Hijacking, but can be one element of the incident so long as additional objectives have also been stated. Hijackings are distinct from Hostage Taking because the target is a vehicle, regardless of whether there are people/passengers in the vehicle. \newline Hostage Taking (Barricade Incident): An act whose primary objective is to take control of hostages for the purpose of achieving a political objective through concessions or through disruption of normal operations. Such attacks are distinguished from kidnapping since the incident occurs and usually plays out at the target location with little or no intention to hold the hostages for an extended period in a separate clandestine location. \newline Hostage Taking (Kidnapping): An act whose primary objective is to take control of hostages for the purpose of achieving a political objective through concessions or through disruption of normal operations. Kidnappings are distinguished from Barricade Incidents (above) in that they involve moving and holding the hostages in another location. \newline Facility/Infrastructure Attack: An act, excluding the use of an explosive, whose primary objective is to cause damage to a non-human target, such as a building, monument, train, pipeline, etc. Such attacks include arson and various forms of sabotage (e.g., sabotaging a train track is a facility/infrastructure attack, even if passengers are killed). Facility/infrastructure attacks can include acts which aim to harm an installation, yet also cause harm to people incidentally (e.g. an arson attack primarily aimed at damaging a building, but causes injuries or fatalities). \newline Unarmed Assault: An attack whose primary objective is to cause physical harm or death directly to human beings by any means other than explosive, firearm, incendiary, or sharp instrument (knife, etc.). Attacks involving chemical, biological or radiological weapons are considered unarmed assaults. \newline Unknown: The attack type cannot be determined from the available information.

For the event, return only a single JSON object with category names as keys and probabilities as values. \newline Example format: \{"Armed Assault": 0.7, "Bombing/Explosion": 0.2, "Unknown": 0.1\}

Event: \newline "<incident text>"

\subsection*{Le Mens and Gallego: manifesto and speech sentences}

The state is the sentence. Each request bundles a Score and a Noul per
dimension: two of each for a manifesto sentence, one of each for a speech
sentence. The Score levels are listed in order and rescaled to 0--100.

\paragraph{Manifestos, economic.} \emph{Score.} ``You will be provided with a
sentence from a party manifesto. Where does this sentence stand on the `left'
to `right' wing scale, in terms of economic policy?'' Levels: Extremely left,
Very left, Left, Somewhat left, Centre, Somewhat right, Right, Very right,
Extremely right. \emph{Noul.} The same opening sentence of instructions, then ``Does this
sentence refer to economic policy?''; true: ``The sentence refers to economic
policy.''; false: ``The sentence does not refer to economic policy.''

\paragraph{Manifestos, social.} As above with ``Where does this sentence stand
on the `liberal' to `conservative' scale, in terms of social policy?'' and
levels from Extremely liberal to Extremely conservative, with ``Neither
liberal nor conservative'' at the centre; the Noul asks about social policy.

\paragraph{Speeches.} The instructions open with their background on the coal
subsidy debate and the coding job verbatim (their \texttt{config\_EU\_debate}),
then ask ``Where does this sentence stand on the `anti-subsidy' to
`pro-subsidy' scale?'', with levels from Extremely anti-subsidy to Extremely
pro-subsidy and ``Neutral'' at the centre. The Noul asks ``Does this sentence
refer to subsidy policy?''. The full text of every question is in
\texttt{data/batch\_inputs/lemens\_prompts\_used.json}.

\subsection*{Weidmann et al.: democracy indicators from country names}
The published prompt (their online Appendix B) with V-Dem v14 codebook wording, one request per country and indicator. The first request sent, for v2clacfree. Instructions:
\begin{quote}\small You are an assistant who evaluates political systems in different countries and years. You will be asked to produce numeric scores derived from your knowledge of this country. The question is as follows: Is there academic freedom and freedom of cultural expression related to political issues?\end{quote}
State: \emph{What is the score for United States of America in 2023?} Score levels, 0 upward:
\begin{enumerate}\setcounter{enumi}{-1}\small
\item Not respected by public authorities. Censorship and intimidation are frequent. Academic activities and cultural expressions are severely restricted or controlled by the government.
\item Weakly respected by public authorities. Academic freedom and freedom of cultural expression are practiced occasionally, but direct criticism of the government is mostly met with repression.
\item Somewhat respected by public authorities. Academic freedom and freedom of cultural expression are practiced routinely, but strong criticism of the government is sometimes met with repression.
\item Mostly respected by public authorities. There are few limitations on academic freedom and freedom of cultural expression, and resulting sanctions tend to be infrequent and soft.
\item Fully respected by public authorities. There are no restrictions on academic freedom or cultural expression.
\end{enumerate}
The published closing line, ``Please only return the score as a number, without any explanation.'', is dropped for JEV and kept for GPT-6 Luna, whose single user message was:
\begin{quote}\small You are an assistant who evaluates political systems in different countries and years. You will be asked to produce numeric scores derived from your knowledge of this country. The question is as follows: Is there academic freedom and freedom of cultural expression related to political issues? The numeric scores are as follows: 0: Not respected by public authorities. Censorship and intimidation are frequent. Academic activities and cultural expressions are severely restricted or controlled by the government. 1: Weakly respected by public authorities. Academic freedom and freedom of cultural expression are practiced occasionally, but direct criticism of the government is mostly met with repression. 2: Somewhat respected by public authorities. Academic freedom and freedom of cultural expression are practiced routinely, but strong criticism of the government is sometimes met with repression. 3: Mostly respected by public authorities. There are few limitations on academic freedom and freedom of cultural expression, and resulting sanctions tend to be infrequent and soft. 4: Fully respected by public authorities. There are no restrictions on academic freedom or cultural expression.\\ What is the score for United States of America in 2023? Please only return the score as a number, without any explanation.\end{quote}

\clearpage
\section*{Appendix B: position bias, transitivity and test-retest}

Only some of these diagnostics need the probability. Position bias, the
choice-based transitivity measures and test-retest agreement on the winner can
be measured from choices alone, and the published article reports the
sparse-triplet measure for its six models (below). Calibration, the
stochastic forms of transitivity and any conditioning on reported confidence
require the probability, which a generative model that exposes token
log-probabilities can also supply, so a typed output does not make any of these
diagnostics possible for the first time. What it changes is that the
probability arrives with every judgment, defined over the named options and at
no extra cost, and
that no answer falls outside the set. Nothing has to be extracted from token probabilities. Whether the probability carries
information is then a question about this model, and the flip rates in Table~\ref{tab:conf} bear
on it. We have no token probabilities for the published generative scorers.
We do for GPT-6 Luna. Averaged over the two orders, its calibration error
against the raters is level with JEV's; in single calls it is higher (0.145
against 0.071). Its interface omits low-probability alternatives, so many of
its probabilities are exactly 0 or 1. Qwen3.8-27B, whose token probabilities we also have, is
better calibrated than JEV on these pairs (0.011 against 0.071 in single calls).

\subsection*{Test-retest}

Replaying \RetestN{} identical pairwise requests reproduced the same winner in
\RetestIdentical\% of cases, with a mean absolute probability shift of
\RetestMeanDp{} and \(r = \RetestR{}\) between passes. Item-level scores, being
continuous, are rarely identical on replay but correlate at \RetestItemR{}.

\subsection*{Position bias}

Asking each pair in both orders identifies the position effect, which a
single-order design cannot. In Study 1 the model favours the second-listed
statement, with a bias of \JevProbAsym{} on the probability scale (\JevChoiceBias{} on the choice scale; negative values favour the second slot); in Study 2 the bias on the probability scale is
\JevProbAsymTwo{}. Flip rates are near-identical across the two studies (8.8\% and 8.5\%) while
the direction of the flips differs, which suggests the effect depends on the
question rather than being a fixed characteristic of the model. TypeSafe reports that the model leans toward
the first option of a Choice. In Study 1 it leans toward the second, so that
documented tendency does not explain the Study 1 result. Consequences for
the recovered scale are small: scores from forward and reversed presentations
correlate at \JevFwdRevR{}. The published models were each run in a single
order, so their departures from an even split confound position bias with any
genuine imbalance between slots, and we do not compare magnitudes across the two
designs.

\subsection*{Transitivity}

On the sparse-triplet measure used in the published article, the model reaches
\JevTransFwd\%, against a published range of \PaperTransLo{} to
\PaperTransHi{}\%. A complete round robin over \JevRRItems{} responses, in which
all \JevTriads{} triads are observable, yields \JevCircular{} circular triads
and a Kendall coefficient of consistence of \JevZeta{}. Over \JevStochTriples{}
ordered triples, weak stochastic transitivity holds in \JevWeakStoch\% of cases
and the strong form in \JevStrongStoch\%, indicating that the model is more
reliable about the direction of a difference than about its magnitude.

\subsection*{Flip rates}

\begin{table}[htbp]
\centering

\small
\begin{tabular}{lrrr}
\toprule
Probability of choice & Pairs & Flip rate & Position bias \\
\midrule
<.50 &  5,537 & 0.3655 & $-$0.1104 \\
.50--.70 &  2,984 & 0.1146 & $-$0.0721 \\
.70--.85 &  2,881 & 0.0139 & $-$0.0263 \\
.85--.95 &  3,862 & 0.0003 & $-$0.0079 \\
>.95 & 12,103 & 0.0000 & $-$0.0005 \\
\bottomrule
\end{tabular}
\normalsize
\caption{Stability and position bias of JEV's pairwise judgments in Study 1, by the probability JEV gives its chosen option (27,367 pairs asked in both orders with no tie). ``Flip rate'' is the share of pairs whose winner changes when the two statements are swapped. ``Position bias'' is the pull toward the first slot on the probability scale; a negative value favours the second.}
\label{tab:conf}
\end{table}

Flip rates decline monotonically across
the five probability bins of Table~\ref{tab:conf}, which cover
\JevDecisivePairs{} decisive pairs.

\clearpage
\section*{Appendix C: the annotation benchmarks in full}

\subsection*{The Gilardi et al.\ relevance benchmark}

The replication archive of \citet{gilardi2023} supplies
the tweets, the instruction they gave ChatGPT, gold labels from two trained
research assistants, and published accuracy figures for ChatGPT and for Amazon
Mechanical Turk.

We use their relevance task, in which a tweet is coded as relating to content
moderation or not, with their instruction verbatim apart from the closing clause
specifying reply format. A Choice question also needs a description of each
answer option, and the two we wrote condense the instruction's own definitions
of relevant and irrelevant tweets, which the instruction states in full. Their gold standard is the label the two assistants
agree on, with disagreements dropped; applying their construction to the archive
reproduces their reported intercoder agreement of \GilHumanIca\% exactly, which
confirms our gold labels match theirs. This yields \GilNTwentyOne{} labelled
tweets in the 2021 sample.

They report two metrics: accuracy against that gold standard, and intercoder
agreement, meaning agreement between two runs at the same temperature, which they report at two temperatures (four runs in all). We therefore ran
JEV four times, as they did, \GilRequests{} requests in total for
\$\GilCost{}.

\begin{table}[htbp]
\centering

\small
\begin{tabular}{lrr}
\toprule
Annotator & Accuracy & Intercoder agreement \\
\midrule
GPT-6 Luna, single run & \LuRelTwentyOne{} & --- \\
JEV 1.13, single run & \GilAccOne{} & --- \\
Qwen3.8-27B, single run & 88.2 & --- \\
JEV 1.13, four runs, unanimity required & \GilAccPen{} & \GilIca{} \\
ChatGPT (temp 1) & \GilChatGPTOne{} & \GilChatGPTIcaOne{} \\
Mechanical Turk & \GilMturk{} & 79.2 \\
ChatGPT (temp 0.2) & \GilChatGPTTwo{} & \GilChatGPTIcaTwo{} \\
Trained annotators & --- & \GilHumanIca{} \\
\bottomrule
\end{tabular}
\normalsize
\caption{Tweet relevance classification, 2021 sample (\GilNTwentyOne{} tweets with agreed gold labels). Rows for ChatGPT, Mechanical Turk and trained annotators are published figures from \citet{gilardi2023}; the others are ours. JEV's intercoder agreement is agreement among its four runs. GPT-6 Luna and the open-weight Qwen3.8-27B were each run once with the published instruction (Appendix H).}
\label{tab:gilardi}
\end{table}

Accuracy is \GilAccOne\% on a single run (Table~\ref{tab:gilardi}). Their
scoring of crowd workers marks a
tweet wrong whenever the two crowd workers who coded it disagree, and applying that rule to our four
runs costs only \GilPenaltyCost{} points, taking accuracy to \GilAccPen\%, because the four runs
are unanimous on \GilUnanimous\% of tweets. Self-consistency is \GilIca\%,
above the \GilHumanIca\% agreement of the two trained annotators, which is the
same pattern \citet{gilardi2023} report for ChatGPT at temperature 0.2. On the smaller 2023 sample
(\GilNTwentyThree{} tweets) accuracy is \GilAccOneTwentyThree\% on a single run
and \GilAccPenTwentyThree\% under unanimity, above every published accuracy on
that sample.

Two qualifications matter more than the margin. The task is easy: a binary
judgment on which two trained humans agree \GilHumanIca\% of the time. A result
of this kind shows only that the model clears a floor of basic competence. And the published comparator is a 2023 model: part of the
\GilMargin{}-point gap between JEV under unanimity and ChatGPT at temperature 1 reflects three years of model development
rather than anything about JEV's constrained output. What the comparison does establish
is that this model, which returns only a constrained decision, handles a
standard annotation task at least as well as the 2023 ChatGPT did, at a
fraction of the cost.

Accuracy alone flatters a classifier when the classes are unbalanced. On relevance, where
\EaGilRelPos\% of the agreed 2021 labels are positive and always answering the
larger class would score \EaGilRelMaj\%, the model's F1 on the relevant class is
\EaGilRelFone{} (precision \EaGilRelPrec{}, recall \EaGilRelRec{}).

The archive does not contain the instructions for the two framing tasks, so we
reconstructed them from the published task definitions. Accuracy there is
\GilProblemAcc\% for problem framing (\GilProblemN{} tweets) and
\GilSolutionAcc\% for solution framing (\GilSolutionN{}), against \EaGilProbMaj\%
and \EaGilSolMaj\% for always answering no, since only \EaGilProbPos\% and
\EaGilSolPos\% of those tweets frame content moderation as a problem or a
solution. F1 on the positive class is \EaGilProbFone{} for problem framing and
\EaGilSolFone{} for solution framing, where the model finds only
\EaGilSolRec{} of the tweets the coders marked. On the 2023 sample, where
\EaGilProbPosTwentyThree\% of \EaGilProbNTwentyThree{} tweets frame content moderation as a
problem, the model's problem-framing accuracy is \EaGilProbAccTwentyThree\%,
below the \EaGilProbMajTwentyThree\% of always answering no (precision
\EaGilProbPrecTwentyThree{}, recall \EaGilProbRecTwentyThree{}). Because the
instructions are ours rather than theirs, none of these figures is comparable
to their published ChatGPT numbers. They show a model that finds most problem
framing and misses most solution framing in the 2021 sample, and that marks far
more problem framing than the coders did in the 2023 sample.

We ran Qwen3.8-27B on the same tweets, once, with the same requests as GPT-6 Luna received and one question per request. Its accuracy on relevance is 88.2\% in 2021 and 79.6\% in 2023. Qwen is level with JEV in 2021 and above it in 2023 (though this lead depends on the seed used for bootstrapping). On problem framing, Qwen is level with JEV in 2021 (both 92.4\%) but above it in 2023 (69.3\% against 61.4\%, a difference of 7.8 points [1.2, 14.5]). On solution framing, Qwen is below JEV in 2021 (81.4\% against 84.7\%, $-3.4$ points [$-4.4$, $-2.3$], and Qwen finds 20\% of the tweets marked by the coders against 41\% for JEV) but level in 2023. Like JEV's figures on framing, Qwen's framing figures are based on our reconstructed instructions, and so are not comparable with the published numbers for ChatGPT either.

\subsection*{The Ornstein et al.\ sentiment benchmark}

For the replication of \citet{ornstein2025}, the \emph{Masterpiece Cakeshop}
preamble is verbatim from their Table 1 and the six few-shot examples per case
are theirs, taken from the \texttt{promptr} package. The classification form of
their \emph{Mazars} preamble sits in their Supplementary Materials, which we could not
retrieve; we took its first sentence verbatim from the version in their Appendix B of the
same instruction and appended the classification sentence from their Table 1.
Correlations use the
\ParN{} of \ParNTotal{} tweets on which every measure is available; pairwise
deletion leaves the ordering unchanged. The intervals below come from \ParBootB{} resamples of those tweets, with
each gap taken as a paired difference within a resample.

With the few-shot examples and positive minus negative, JEV sits \ParGapFourFew{}
[\ParGapFourFewLo{}, \ParGapFourFewHi{}] below GPT-4 and \ParGapNlpFew{}
[\ParGapNlpFewLo{}, \ParGapNlpFewHi{}] above the RoBERTa model, and under the
principal component it stays above that model by \ParGapNlpPcaFew{}
[\ParGapNlpPcaFewLo{}, \ParGapNlpPcaFewHi{}]. Without the examples it cannot be
told apart from the RoBERTa model under positive minus negative
(\ParGapNlpZero{} [\ParGapNlpZeroLo{}, \ParGapNlpZeroHi{}]), and under the
principal component it falls \ParGapNlpPcaZero{} [\ParGapNlpPcaZeroLo{},
\ParGapNlpPcaZeroHi{}] below it.

With the few-shot examples, and with the GPT-4 rule applied to its twenty most probable first tokens, Qwen3.8-27B scores 0.841 [0.811, 0.868], which is 0.052 [0.025, 0.078] higher than GPT-4 and 0.182 [0.141, 0.226] higher than the RoBERTa model. Without examples, Qwen3.8-27B produced a prose answer (rather than a label) for 941 out of 945 tweets. Its zero-shot score of 0.668 is therefore based on the first token of a prose answer, and is not comparable with the other scores (Table~\ref{tab:parrot}).

\begin{table}[htbp]
\centering

\small
\begin{tabular}{lllr}
\toprule
Measure & Kind & Scoring rule & $\rho$ with experts \\
\midrule
Qwen3.8-27B (few-shot) & generative & positive $-$ negative & 0.841 \\
GPT-6 Luna (few-shot) & generative & positive $-$ negative & 0.816 \\
GPT-3 (few-shot) & generative & principal component & 0.807 \\
GPT-6 Luna (zero-shot) & generative & positive $-$ negative & 0.791 \\
GPT-4 (few-shot) & generative & positive $-$ negative & 0.789 \\
JEV (few-shot) & constrained output & positive $-$ negative & 0.740 \\
JEV (few-shot) & constrained output & principal component & 0.714 \\
JEV (zero-shot) & constrained output & positive $-$ negative & 0.701 \\
Qwen3.8-27B (zero-shot)$^{a}$ & generative & positive $-$ negative & 0.668 \\
TweetNLP RoBERTa & fine-tuned encoder & principal component & 0.659 \\
JEV (zero-shot) & constrained output & principal component & 0.604 \\
Naive Bayes & supervised & P(positive) & 0.291 \\
\bottomrule
\multicolumn{4}{l}{\scriptsize $^{a}$~941 of 945 replies were prose, not a label.}
\end{tabular}
\normalsize
\caption{Twitter sentiment, replicating Application 1 of \citet{ornstein2025}: correlation with the mean of three expert ratings across the 907 tweets on which every measure is available. Their four measures are recomputed from their files with their code and reproduce their published figures. Each is scored by the rule their figure code applies: the first principal component of the three class probabilities for GPT-3 and TweetNLP (the rule they pre-registered), positive minus negative for GPT-4. JEV is shown under both rules; GPT-6 Luna and the open-weight Qwen3.8-27B under GPT-4's (Appendix H). Without the examples Qwen answered 941 of 945 tweets in prose, so its zero-shot score is not comparable.}
\label{tab:parrot}
\end{table}

\clearpage
\section*{Appendix D: A/B correctness}

Correlation between two latent scales can hide disagreement about the individual
comparisons that produced them. Table~\ref{tab:abacc} therefore scores each
scorer on the decision it actually makes. Column 3 is accuracy on the
expert-judged pairs, the share of the 460 judgments matched; the main text reports F1 on the same pairs. Column 4 is agreement with the majority of
the other scorers (the six published models and JEV) across all 28,040 published pairs, which is a far larger
sample but measures agreement with the other models rather than correctness.

Nine further models from an unpublished extension by the same authors drew their own
pairs, sharing only 782 of the 28,040, so they are not in the table; their scores are in the replication archive. GPT-6 Luna and Qwen3.8-27B did not judge the full published grid either, so they cannot enter column 4; their expert-pair accuracy (0.780 and 0.787) and F1 are in Table~\ref{tab:f1}.

\begin{table}[htbp]
\centering

\small
\begin{tabular}{llrr}
\toprule
Scorer & Mode & Expert pairs & Other models' majority \\
\midrule
Llama 3.1 405B & pairwise & 0.815 & 0.949 \\
Gemma 3 27B & pairwise & 0.813 & 0.928 \\
JEV & pairwise & 0.796 & 0.910 \\
GPT-4o & pairwise & 0.817 & 0.906 \\
GPT-4o mini & pairwise & 0.824 & 0.899 \\
Gemma 3 4B & pairwise & 0.739 & 0.875 \\
Llama 3.1 8B & pairwise & 0.737 & 0.870 \\
\bottomrule
\end{tabular}
\normalsize
\caption{A/B correctness rather than correlation. Column 3 is accuracy on
the expert-judged pairs; column 4 is agreement with the majority of
the other six scorers across all 28,040 published pairs. JEV's accuracy in column 3 is averaged over both presentation orders (0.800 in the published order alone; Table~\ref{tab:f1}). Nine further models from an unpublished extension by the same authors drew their own pairs and share only 782 of
those pairs, so they are not shown.}
\label{tab:abacc}
\end{table}

\clearpage
\section*{Appendix E: what is known about the model}

We requested JEV 1.13 as \texttt{typesafe/jev-1.13} through OpenRouter on 19, 22
and 23 September 2026; every response reported \texttt{typesafe/jev-1.13-20260917},
which we take to be TypeSafe's \texttt{jev-1.13.0}. The commercial term is ``System One''. The vendor
documents price, throughput and nine failure modes but publishes no
public-benchmark results or calibration metrics \citep{typesafe2026}. This
appendix sets out what the vendor states, what the outputs reveal and what
remains unknown.

\subsection*{What the vendor states}

TypeSafe says it built JEV with ``a new architecture, a new sampler, and a new
training algorithm'' \citep{typesafe2026}. It describes none of the three
technically. It calls the sampler parallel, producing every output in one query
instead of token by token, and says JEV writes no text and is not a chat or
code-completion LLM. It describes its training method, Reinforcement Learning for
Calibrated Decisions (RLCD), only by its aim, calibrated probabilities, and
gives no formal objective for it. TypeSafe says it makes its own training data,
and its chief executive told TechCrunch that the data are exclusively synthetic
while saying little about the architecture \citep{fernholz2026}. That account would
not rule out that the model has seen our benchmarks, since text another model
writes carries what that model learned; a benchmark built after the version
date would settle it. TypeSafe's primer presents RLCD as a third way,
after reinforcement learning from human feedback and from verifiable rewards, to
post-train pretrained language models, which we read as implying an unnamed
pretrained base. The same weights serve every customer.

The documentation says the state is read once and every question is evaluated
against it in parallel and in isolation, with each Score level judged against
the state on its own wording, without its number or neighbours. It offers no determinism guarantee. The API takes no temperature or
seed.

TypeSafe lists nine known failure modes for version 1.13 (page last reviewed
2 October 2026). Among them, it says, the model reads questions literally,
handles double negatives and multi-hop questions less reliably, counts poorly,
loses accuracy as irrelevant text fills the state, and leans toward the first
option of a Choice. It also warns that the expected Score should not be read
as an exact number between two levels.

Several of these map onto our prompts. Study 2 quotes a survey item containing
a negation, keeps two errors from the published wording in its pairwise
options, and asks which answer is more uncertain while its Score runs from
uncertain to certain. In both studies, every pairwise state also holds two
statements where a pointwise state holds one. The archive allows only indirect
checks. Study 2's judgments are no less stable than Study 1's: its flip rate is
\JevFlipRateTwo\% against \JevFlipRate\%, and its position effect is the
smaller (Appendix B). The uncertain arm raises the pairwise scale
(\JevUncertEffect{}) and lowers the certainty-worded Score
(\JevUncertScoreEffect{} [\JevUncertScoreLo{}, \JevUncertScoreHi{}]), so the
two polarities point the same way on the uncertain-arm effect. Neither check shows
that the failure modes are absent. Rerunning Study 2 with the errors corrected
and the Score's polarity reversed, and rerunning a sample of Study 1 pairs with a neutral
paragraph added to the state, would test them directly; we have not done so.

\subsection*{What the outputs reveal}

Everything below comes from the \FxTotalOk{} successful responses of the first
four applications, collected on \FxDateFirst{} and \FxDateLast{}, and their run
logs.
We flag each inference about how JEV is built. Although no response carries
generated text, each reports an output-token count, constant within each of our
\FxOutTokSchemas{} question schemas and running from \FxOutTokDileo{} for party
comparisons to \FxOutTokGilardi{} for three bundled Gilardi questions. We infer
that the schema fixes the count and the answer never moves it. That fits the
vendor's statement that JEV writes no text. Only input tokens are billed, so the
count may be nominal.

Every probability is a multiple of 0.01, and \FxProbZeroShare\% of Choice and
Score probabilities are exactly zero, so we infer rounding to the nearest
hundredth: a zero stands for a value below 0.005. Probability vectors sum to
1.00 in \FxSumOneShare\% of cases and to 0.99 in the other \FxSumShortN{}, never
above one, which we read as a sum-to-one adjustment after rounding. The returned
choice is the most probable option in \FxArgmaxShare\% of \FxChoiceN{} Choice
answers, as documented. The Score is the documented expected level counted from
zero. Allowing for rounding, that expectation reproduces \FxScoreFeasible\% of
\FxScoreN{} returned Scores, and an expectation counted from one never matches exactly. The
rescaling of JEV's Scores to the 0--10 range in the survey application relies on this.

TypeSafe now documents how confidence is computed (accessed 4 October 2026):
for a Choice with $n$ options it is $(p_{\max} - 1/n)/(1 - 1/n)$, a rescaling of
the top probability. On our two- and three-option Choice questions, confidence
behaves as \FxConfFormula{}, which matches it. On the \FxConfPreN{} Choice answers in which one probability carries a floating-point remainder from the sum-to-one adjustment, so that its value before the adjustment can be recovered, the formula reproduces the returned confidence within 0.01
in \FxConfPreWithin\% of cases from the pre-adjustment top probability and in
\FxConfPostWithin\% from the returned one, so we infer confidence is computed
before the adjustment. For a Score, TypeSafe gives one minus the probability-weighted distance from
the most likely level, divided by the mean absolute deviation of an even spread
over the levels, floored at zero. Applied to the rounded probabilities returned,
it reproduces the returned confidence to within 0.02 in 96\% of our five-level
and 86\% of our ten-level survey Scores.

Answers vary slightly on replay. Across \FxRetestChoiceN{} byte-identical
pairwise replays from both studies, \FxRetestSameChoice\% returned the same
choice (Appendix B covers Study 1 alone), but the probability vector was
identical in only \FxRetestSameVector\%, and each option's probability moved by
\FxRetestMeanAbsDiff{} on average. Asking the same questions together or one at a time,
seconds apart (\FxBundleN{} pairs), moved Choice probabilities by
\FxBundleMeanAbsDp{} on average, between the shifts for duplicates within one
run (\FxDupMeanAbsDiff{}) and for replays a median \FxRetestGapMinutes{} minutes
later (\FxRetestMeanAbsDiff{}). We read this as bounding any bundling effect at
about the size of replay variation, as the vendor's account predicts.

\subsection*{What remains unknown}

We do not know the architecture (encoder, decoder, diffusion or otherwise),
whether a request is one forward pass, the parameter count, or which base or
teacher model, if any, JEV derives from. Nor do we know how options and levels
are scored, what an output token counts, or whether replay variation arises in
the model or the serving stack. TypeSafe publishes no model changelog. We
therefore cannot tell whether the weights changed between our collection dates.

Nothing in the outputs separates a new architecture from a pretrained or
distilled language model whose scores over the answer set are read off,
rounded and renormalised. Rounding to hundredths, the sum-to-one adjustment, a
confidence computed from the top probability, schema-fixed output-token counts
and small replay variation are properties of an interface, and a language model
behind a constrained output head could produce every one of them. A technical report, open weights
or a named base model would settle the question. Short of that, two checks on
the archive would bear on it without settling it: whether the input-token
counts the endpoint reports match one tokenizer family exactly, which would
point to a base model from that family if the vendor counts with the model's
own tokenizer, and whether the model's disagreements with other scorers coincide
with one generative model's more often than shared difficulty explains, which
would point to a teacher model. We have run neither.

\clearpage
\section*{Appendix F: the conflict benchmark in full}

\subsection*{Test set}

\citet{brandt2026} evaluate on GTD incidents from 2017 to 2020. Their posted
code, applied to the 0522 GTD release, yields 38,157 incidents, while their archived
predictions cover \BrtN{}. The difference is one filter that neither the paper
nor the code states: dropping the 448 incidents whose summary does not open
with a complete calendar date, 443 of them dated to a month only. With that
filter our rebuilt test set matches their event identifiers exactly. The text
each model sees is the summary followed by the motive, as their code builds it.

\subsection*{Reproducing the published figures}

Scoring their per-incident predictions with our reimplementation of their
parser and metrics reproduces all \BrtGateCells{} cells of their Tables 2, 5 and
6 within rounding. Their macro averages are unweighted means over the nine
types, with a type never predicted scoring zero. A footnote describes them as
support-weighted, but only the unweighted mean reproduces the printed values.
Every model in Table~\ref{tab:brandt} is scored by that rule on the same
\BrtN{} incidents, with any answer outside the nine types counted wrong. Qwen3.5-9B is parsed exactly as their generative models were, taking the
highest-probability key of the returned JSON object.

\subsection*{The runs and the prompted baselines}

Qwen3.5-9B ran through OpenRouter, pinned to a single provider's bf16 deployment
with no fallback, reasoning disabled, JSON output and a fixed seed (Appendix~A). All \BrtN{} responses in each arm came from that deployment. The original
authors ran mid-2024 versions of their generative models locally at 4-bit
precision \citep{brandt2026}.

This design cannot say how much of Qwen3.5-9B's lead over the published prompted
models reflects their age and how much their precision or serving. The same
holds for JEV's lead over the published prompted models. Its lead over Qwen3.5-9B,
run at the same time through the same router, is the cleaner comparison, though
the two arms still differ in prompt and interface. Running at 4-bit is costly
on this task even for a fine-tuned model: ConflLlama reaches macro F1
\BrtCllQfourFone{} at 4-bit against \BrtCllQeightFone{} at 8-bit. Qwen3.5-9B is
one model of 9 billion parameters, and JEV's size is unknown (Appendix E).

On 30 September 2026 we ran Qwen3.8-27B through OpenRouter, pinned to a single host (Parasail) running it at 8-bit precision, with reasoning turned off and with the request messages that GPT-6 Luna received. When we give it the category names, it has an accuracy of 0.680 and a macro F1 of 0.518. Both are lower than JEV's, by 0.034 [0.032, 0.037] in accuracy and 0.038 [0.026, 0.049] in macro F1. When we give it the codebook instead, it has an accuracy of 0.689 and a macro F1 of 0.593 (Table~\ref{tab:brandt}), an improvement of 0.009 [0.005, 0.012] in accuracy and 0.075 [0.067, 0.083] in macro F1. Both improvements are level with JEV's (differences in differences, JEV's gain minus Qwen's, of $-0.003$ [$-0.006$, 0.001] for accuracy and 0.008 [$-0.004$, 0.020] for macro F1), but smaller than GPT-6 Luna's (0.035 and 0.090).

\subsection*{Ties and decoding}

JEV reports probabilities to the hundredth, so ties occur. For the nine-way
question the prediction is the option JEV returns, whose probability is tied at
the top in \BrtChoiceTies{} incidents. With nine separate questions the highest
probability is tied in \BrtOvrTies{} incidents; we break those ties at random
with a fixed seed, and breaking them by the codebook hierarchy instead moves
accuracy by less than 0.004. Under hierarchy decoding we take the highest-ranked type whose probability reaches 0.5,
and the maximum when none does.

Where only one of the nine-way question and the nine separate questions
is right, it is the single question for \BrtOvrOnlyChoice{} incidents and the
separate questions for \BrtOvrOnlyOvr{}.

\subsection*{Recall by type}

\begin{table}[htbp]
\centering

\scriptsize%
\setlength{\tabcolsep}{2pt}
\begin{tabular}{>{\raggedright\arraybackslash}p{2.0cm}rrrrrrrrrrr}%
\toprule
Attack type & Support & \shortstack{Confli-\\BERT} & \shortstack{ConflLlama\\8-bit} & \shortstack{JEV\\names} & \shortstack{JEV\\codebook} & \shortstack{Qwen 3.5\\names} & \shortstack{Qwen 3.5\\codebook} & \shortstack{GPT-6\\Luna\\names} & \shortstack{GPT-6\\Luna\\codebook} & \shortstack{Qwen\\27B\\names} & \shortstack{Qwen\\27B\\codebook} \\
\midrule
Assassination & 2,990 & 0.74 & 0.59 & 0.29 & 0.54 & 0.02 & 0.38 & 0.48 & 0.66 & 0.28 & 0.75 \\
Armed Assault & 9,079 & 0.81 & 0.75 & 0.90 & 0.94 & 0.94 & 0.89 & 0.79 & 0.89 & 0.88 & 0.91 \\
Bombing/\allowbreak Explosion & 14,508 & 0.97 & 0.92 & 0.91 & 0.81 & 0.84 & 0.65 & 0.92 & 0.86 & 0.85 & 0.73 \\
Hijacking & 154 & 0.64 & 0.54 & 0.60 & 0.70 & 0.44 & 0.58 & 0.38 & 0.50 & 0.55 & 0.56 \\
Hostage taking, barricade & 230 & 0.38 & 0.27 & 0.33 & 0.46 & 0.26 & 0.37 & 0.23 & 0.46 & 0.29 & 0.41 \\
Hostage taking, kidnapping & 3,495 & 0.88 & 0.86 & 0.75 & 0.82 & 0.77 & 0.81 & 0.71 & 0.81 & 0.69 & 0.71 \\
Facility/\allowbreak infrastructure & 2,624 & 0.85 & 0.74 & 0.72 & 0.79 & 0.33 & 0.72 & 0.79 & 0.87 & 0.69 & 0.78 \\
Unarmed Assault & 301 & 0.62 & 0.50 & 0.25 & 0.44 & 0.18 & 0.31 & 0.35 & 0.40 & 0.18 & 0.31 \\
Unknown & 4,328 & 0.55 & 0.39 & 0.00 & 0.01 & 0.03 & 0.07 & 0.00 & 0.03 & 0.00 & 0.01 \\
\bottomrule
\end{tabular}
\normalsize
\caption{Recall by attack type on the \BrtN{} test incidents. Support is the
number of incidents of that type. ConfliBERT and ConflLlama are recomputed from
Brandt et al.'s per-incident predictions. GPT-6 Luna is a current generative
model (Appendix H). Qwen 3.5 is Qwen3.5-9B and Qwen 27B is the open-weight Qwen3.8-27B. ``Names'' columns give the model only the category names; ``codebook'' columns add the GTD codebook definitions.}
\label{tab:brandtrecall}
\end{table}

Table~\ref{tab:brandtrecall} shows where the codebook helps. For JEV it raises
recall on eight of the nine types, Unknown only trivially, and lowers it on
Bombing/Explosion. That fall is consistent with the coding hierarchy, which
ranks Bombing/Explosion below four other types.

GTD reserves Assassination for
attacks on specific, prominent individuals, and JEV's recall on it rises from
\BrtJevBareAssRecall{} to \BrtJevCbAssRecall{} (Table~\ref{tab:brandtrecall}). The codebook's hierarchy also fixes the gold label. Of the test
incidents, \EaGtdMultiShare\% carry more than one recorded type, and in all
\EaGtdMultiN{} of them the first, the label every model is scored on, is the
type the hierarchy ranks highest. A model given only the names is not told that
rule, so on those incidents an answer naming another recorded type, such as
Bombing/Explosion for an assassination by bomb, counts as wrong.

Macro
F1 weights the nine types equally and accuracy weights incidents, and for both JEV and Qwen3.5-9B the codebook raised recall on rare and narrowly defined types while
lowering it on Bombing/Explosion, the most common (Table~\ref{tab:brandtrecall}). Qwen3.5-9B also started lower, recalling almost no
Assassinations from the names alone (\BrtQwenBareAssRecall{}), so it had more to
gain in macro F1, which scores a type a model seldom predicts near zero.
For Qwen3.8-27B too, the codebook increases recall for Assassination (0.28 to 0.75), which is the biggest increase in Table~\ref{tab:brandtrecall}. It decreases recall for Bombing/Explosion (0.85 to 0.73).

\subsection*{Unknown}

JEV almost never answers Unknown: it gives that label to \BrtJevBarePredUnk\% of
incidents from the names alone and to \BrtJevCbPredUnk\% when the codebook
defines it. A model reproducing memorised codes would give Unknown as readily as
any other, so this is also weak evidence against JEV having memorised the test set's
codes, public in GTD releases before its version date, though not against
partial exposure. Comparisons restricted to incidents of known type are also not neutral: the
fine-tuned models' Unknown answers on the incidents kept still count as errors. Qwen3.5-9B, which writes its answer, also
seldom answers Unknown even with the codebook (recall \BrtQwenCbUnkRecall{},
against \BrtJevCbUnkRecall{} for JEV), so most of the shortfall is shared by a
generative model given the same definition. Qwen3.8-27B also has a recall for Unknown of 0.000 from the names and 0.006 with the codebook.

Whatever part of the fine-tuned models' lead the codebook leaves is credited
to examples only by elimination. It includes base rates, conventions the
codebook does not spell out and ConfliBERT's pretraining on conflict text.
The Unknown result also suggests that what a code means in practice is partly
learned from examples, so the design cannot divide the fine-tuned models' lead
between examples and meaning.

\subsection*{Multi-label classification}

The GTD records up to three attack types per incident. Scored as a multi-label
task by their rule, which keeps every type with a probability of at least 0.5,
the nine-way question reaches a subset accuracy of \BrtMlChoiceSubset\% (category names only; 67.5\% with the codebook),
against \BrtMlCbertSubset\% for ConfliBERT and \BrtMlQeightSubset\% for the
8-bit ConflLlama. Its probabilities sum to one, so that rule gives it at most
one type per incident, and since most incidents carry a single type, this costs
little. The nine separate questions reach \BrtMlOvrSubset\%, because they
say yes too readily: on average \BrtMlOvrCard{} of the eight known types clear
0.5 per incident, where coders record \BrtMlTrueCard{}. With its written probabilities, Qwen3.8-27B reaches a subset accuracy of 63.7\% from the category names and 65.2\% with the codebook, lower than JEV. GPT-6 Luna's written probabilities give 66.6\% and 70.1\%. Neither model wrote two types at 0.5 or above for any incident, so both are effectively single-label.

\subsection*{The BBC task}

Their Table 2 also classifies \BrtBbcN{} BBC news articles as conflict or not.
JEV's F1 on the conflict class is \BrtBbcJevFone{}, level with Llama 3.1 8B
(\BrtBbcLlamaFone{}) and above Gemma 2 9B (\BrtBbcGemmaFone{}), and well below
ConfliBERT (\BrtBbcCbertFone{}). Its accuracy is \BrtBbcJevAcc{}.
With the prompt as published, both current generative models mark very few articles as conflict. GPT-6 Luna marks 19 articles as conflict, 5 of them among the 53 conflict articles (conflict F1 0.139). Qwen3.8-27B marks 15, 6 of them conflict articles (conflict F1 0.176). Both are lower than JEV (GPT-6 Luna minus JEV $-0.182$ [$-0.308$, $-0.050$] and Qwen3.8-27B minus JEV $-0.145$ [$-0.278$, $-0.017$]). Since most articles are not about conflict, Qwen3.8-27B's accuracy of 0.826 is still level with JEV's ($-0.003$ [$-0.028$, 0.022]).

\clearpage
\section*{Appendix G: further results and checks}

\subsection*{Party positions}

\citet{dileo2025} run each comparison seven times and keep the modal answer,
reporting 93.23\% congruence across repeats; we run a single pass. Modal voting
reduces run-to-run noise, but on the survey comparisons JEV returns the
same winner on \RetestIdentical\% of replays (Appendix B), so we expect seven
passes would change few of its decisions here either, though we have no
replays of the party comparisons.

Their scores are not in the archive and their figures are
transcribed to two decimals, so only our correlations carry intervals
(Table~\ref{tab:dileo}). The voter comparison rests on \DilTevNLo{} to
\DilTevNHi{} parties a year. Our CHES correlation is stable, running
\DilChesLo{} to \DilChesHi{} across the \DilChesN{} elections with a CHES wave,
which span four decades of European party systems. \citet{dileo2025} attribute
the lower agreement with manifesto scores, which we reproduce, to the
sensitivity of manifesto scaling.
In panel b of Table~\ref{tab:dileo} we add Qwen3.8-27B and a fourth benchmark, voters' placements in the CSES, which are available for 1999 to 2014 (between 95 and 142 parties a year). The correlations with this benchmark are between 0.76 and 0.84 for JEV's scale, between 0.70 and 0.78 for GPT-6 Luna's and between 0.58 and 0.67 for Qwen's, so the ordering is the same as with the experts. These are single passes, without paired intervals.

Because every pair here is presented in one order, as in their design, the
first-listed party is chosen \DilFirstShare\% of the time. A single-order study
cannot distinguish that departure from genuine imbalance between slots, and
here the slots are not exchangeable: the published duplet files fix each pair's
order by country and then by party name, so a preference for one slot, of the
kind Study 1 shows (Appendix B), would move parties listed early in that order
relative to those listed late rather than average out.

Nor can this design exclude contamination. The model sees nothing but each
party's name and country, so it answers from what it has stored about the
parties, and the expert, manifesto and voter placements it is validated against
were public before the model's version date, and are listed beside each party's name in
\citeauthor{dileo2025}'s own files. A model that had learned those placements would also
agree most closely with the experts, so the ordering across benchmarks cannot
separate recall of the benchmark from knowledge of the parties. Being newer,
JEV has also had more chance to see the placements than GPT-3.5 had, and part
of its margin over the published figures may reflect that exposure.

Splitting the parties by coverage bears on this without settling it. A model
recalling the expert placements should place the parties the expert survey covers
better than those it leaves out, and since only covered parties have a CHES
placement, we compare the two groups on the manifesto scores, which most
parties in both carry. Over the \DilCovNYears{} years with a CHES wave, JEV's
scale correlates with them at a mean of \DilCovRCov{} [\DilCovRCovLo{},
\DilCovRCovHi{}] among parties CHES places that year and \DilCovRUnc{}
[\DilCovRUncLo{}, \DilCovRUncHi{}] among the rest, a gap of \DilCovD{}
[\DilCovDLo{}, \DilCovDHi{}] in a bootstrap over parties that holds JEV's
scores fixed, positive in \DilCovNPos{} of the \DilCovNYears{} years. That is
the pattern recall predicts. But \DilCovPYUncCountry{} of the \DilCovPYUnc{}
party-years without a placement belong to countries CHES did not survey that
year, and a model that knew less about those countries' parties would show the
same gap; the manifesto scores spread about as widely in both groups (standard
deviations averaging \DilCovSdCov{} and \DilCovSdUnc{} points), so a narrower
range does not explain it. The split cannot tell recall from salience, and a
model could also have recalled the manifesto scores themselves.

\subsection*{The survey application}

\begin{table}[htbp]
\centering

\small
\begin{tabular}{lrrrrrr}
\toprule
Measure & N & Precision & Recall & F1 & 95\% CI & Accuracy \\
\midrule
GPT-4o mini (paper) & 460 & 0.835 & 0.845 & 0.840 & [0.797, 0.876] & 0.824 \\
Gemma 3 27B (paper) & 460 & 0.821 & 0.841 & 0.831 & [0.787, 0.869] & 0.813 \\
GPT-4o (paper) & 460 & 0.852 & 0.805 & 0.828 & [0.780, 0.869] & 0.817 \\
JEV (Noul) & 460 & 0.812 & 0.829 & 0.821 & [0.772, 0.858] & 0.802 \\
Llama 3.1 405B (paper) & 459 & 0.877 & 0.769 & 0.820 & [0.770, 0.859] & 0.815 \\
JEV (rev) & 460 & 0.799 & 0.841 & 0.819 & [0.773, 0.858] & 0.798 \\
GPT-6 Luna (fwd) & 460 & 0.748 & 0.900 & 0.817 & [0.772, 0.856] & 0.780 \\
JEV (order-averaged) & 460 & 0.803 & 0.829 & 0.816 & [0.769, 0.856] & 0.796 \\
JEV (fwd) & 460 & 0.827 & 0.801 & 0.814 & [0.764, 0.852] & 0.800 \\
Qwen3.8-27B (fwd) & 460 & 0.884 & 0.701 & 0.782 & [0.721, 0.831] & 0.787 \\
Gemma 3 4B (paper) & 460 & 0.796 & 0.701 & 0.746 & [0.685, 0.799] & 0.739 \\
Llama 3.1 8B (paper) & 460 & 0.849 & 0.629 & 0.723 & [0.656, 0.781] & 0.737 \\
\bottomrule
\end{tabular}
\normalsize
\caption{Agreement with expert raters: 460 judgments of 243 pairs. F1, defined as in the published article, treats ``human picked statement 1'' as the positive class, so it depends on order; accuracy is also shown. ``fwd'' is the published presentation order, ``rev'' the reverse and ``(paper)'' marks the published scorers. Intervals: 2,000 bootstrap resamples of pairs; rater clustering is not modelled. GPT-6 Luna and the open-weight Qwen3.8-27B are ours, in the published order (Appendix H).}
\label{tab:f1}
\end{table}

Bootstrapping the \HumanNPairs{} judged pairs gives F1 intervals
\FoneCIWidthLo{} to \FoneCIWidthHi{} wide (Table~\ref{tab:f1}). The paired
difference from every generative scorer with a higher F1 has an interval that includes zero, even
for GPT-4o mini, which leads by \JevFoneGapTop{} [\JevFoneGapTopLo{},
\JevFoneGapTopHi{}]. 
In the published order, Qwen3.8-27B (0.782 [0.721, 0.831]) is level with both JEV ($-0.032$ [$-0.070$, 0.003]) and GPT-6 Luna ($-0.035$ [$-0.086$, 0.012]).

Two features of the benchmark bound what it can show. JEV was asked
about these pairs with the published human-comparison prompt, which has no
``equal'' option because the raters had none, so the specification scored here
differs from the grid's by that option. And the raters disagree with one
another. On the \HcNMulti{} pairs judged more than once (\HcNJudgMulti{}
judgments), two judgments of the same pair pick the same statement \HcAgree\% of
the time. No scorer could agree with more than \HcCeiling\% of those judgments,
and JEV agrees with \HcJevAcc\%. A scorer that follows the raters'
majority agrees with any one of them more often than two of them agree with
each other, so differences of a few points between the rows of
Table~\ref{tab:f1} are small against the raters' own disagreement.

The Study 2 regression also compares the
control arm with the certain arm. There JEV's pairwise scale gives \JevControlEffect{}
[\JevControlLo{}, \JevControlHi{}], an interval that includes zero, so it
recovers the larger of the two contrasts and not the smaller. Among the six
published models, \PaperControlNPos{} separate those two arms with the control
arm above and \PaperControlNNeg{} with it below. 

\clearpage
\section*{Appendix H: a current generative comparator}
\label{app:gpt6luna}

The published prompted models are the ones the original authors used,
from 2023 to mid-2024. To set JEV beside a current one, we ran OpenAI's GPT-6
Luna on every test through its Batch API on 25 September 2026 (on 28 September for the manifesto and speech sentences and, as described below, the democracy indicators). Each paper's
published prompt was sent verbatim, reply-format lines included, with
temperature 0, a fixed seed and reasoning effort set to none. The one change
the interface forced is in the sentiment task, where \citeauthor{ornstein2025}'s request
for a single token with twenty alternatives became a short reply with the
five alternatives the interface allows; their score reads the first token
either way. The conflict prompts are, byte for byte, those Qwen3.5-9B received.
Tests with more than 3,000 items were piloted first (by respondent in the
survey, within 2009 for party positions, stratified by attack type for the
GTD) before the full run.

We compare GPT-6 Luna with JEV on eight checks. Each is scored by the original paper's
rule, through code that first reproduced JEV's reported figures exactly, and
the difference from JEV is read from a paired bootstrap.
Table~\ref{tab:gpt6luna} gives the differences from JEV and the reading of
each. The scores themselves stand beside JEV's in
Tables~\ref{tab:brandt}, \ref{tab:vdem}, \ref{tab:dileo} and~\ref{tab:lemens} and in the
appendix tables of the same tests (Tables~\ref{tab:gilardi}, \ref{tab:parrot},
\ref{tab:brandtrecall} and~\ref{tab:f1}). Those tables also hold cells outside the
eight checks, such as GPT-6 Luna's correlations with the voter and manifesto
benchmarks. They are descriptive, and no claim rests on them. On the two
reconstructed framing tasks, which Table~\ref{tab:gpt6luna} omits, GPT-6 Luna is
level with JEV in all four task-by-sample cells. With the codebook, it reaches accuracy \LuBrtCbAcc{} and
macro F1 \LuBrtCbFone{} on the GTD incidents, above JEV's codebook arm in
accuracy and level in macro F1. The \LuRequests{} requests cost
\$\LuCost{}. Across \LuCostTasks{} matched tasks, GPT-6 Luna's price per decision was
\LuCostRatioLo{} to \LuCostRatioHi{} times JEV's, and at most
\LuCostRatioCwHi{} times if every uncached input token is billed at the
posted cache-write rate.

\begin{table}[htbp]
\centering

\footnotesize
\setlength{\tabcolsep}{3pt}
\begin{tabular}{>{\raggedright\arraybackslash}p{3.4cm}>{\raggedright\arraybackslash}p{2.9cm}llll}
\toprule
Test & Statistic & GPT-6 Luna & JEV & Difference & Reading \\
\midrule
Tweet relevance, 2021 & accuracy (\%) & 91.1 & 89.0 & 2.1 [0.7, 3.5] & above \\
Tweet relevance, 2023 & accuracy (\%) & 84.2 & 76.2 & 8.0 [4.1, 11.9] & above \\
Tweet sentiment, few-shot & $\rho$ with experts & 0.816 & 0.740 & 0.076 [0.039, 0.112] & above \\
GTD attack type, names & accuracy & 0.709 & 0.715 & $-$0.006 [$-$0.009, $-$0.003] & below \\
GTD attack type, names & macro F1 & 0.544 & 0.555 & $-$0.011 [$-$0.025, 0.002] & level \\
BBC conflict & conflict F1 & 0.139 & 0.321 & $-$0.182 [$-$0.308, $-$0.050] & below \\
Party positions & CHES $r$, mean of 8 years & 0.844 & 0.877 & $-$0.051 to $-$0.017 & level in 8 of 8 \\
Survey, expert pairs & F1 & 0.817 & 0.814 & 0.004 [$-$0.031, 0.043] & level \\
Survey, Study 1 scale & $r$ with GPT-4o's scale & 0.935 & 0.937 & $-$0.002 [$-$0.008, 0.004] & level \\
Survey, calibration & error vs raters & 0.114 & 0.109 & 0.005 [$-$0.031, 0.040] & level \\
\midrule
Democracy indicators & mean country $r$ with V-Dem & 0.628 & 0.680 & $-$0.052 [$-$0.068, $-$0.037] & below \\
Democracy indicators & exact agreement & 0.516 & 0.543 & $-$0.027 [$-$0.038, $-$0.016] & below \\
Democracy indicators & calibration error & 0.390 & 0.057 & 0.333 [0.318, 0.347] & above \\
\bottomrule
\end{tabular}
\normalsize
\caption{GPT-6 Luna against JEV on every test, under each paper's published prompt (one pass, temperature 0, reasoning off; Appendix H). Differences are GPT-6 Luna minus JEV's primary specification, with paired bootstrap 95\% intervals; above, level and below say whether the interval lies above zero, contains it or lies below it. For calibration error lower is better, so ``above'' there favours JEV. The last three rows (democracy indicators) were added later.}
\label{tab:gpt6luna}
\end{table}

Three further observations stand out. First, GPT-6
Luna is not deterministic. Requests sent twice with identical bodies, once in a
pilot and again in the full run, returned the same answer \LuRepeatLo\% to
\LuRepeatHi\% of the time, against \RetestIdentical\% for JEV's replays
(Appendix B). No interval in Table~\ref{tab:gpt6luna} includes that variation.
Second, it has position effects of its own. On the expert-judged survey pairs
it chose the statement shown first more often than the raters did, and on the
party duplets it chose the party listed second more often than the expert
placements imply. Third, its interface omits low-probability alternatives, so
many of its token probabilities are exactly 0, 0.5 or 1, and a score built from
them is coarser than GPT-4's. The scripts, request files and every response are
in the replication archive.

The democracy indicators were added later. GPT-6 Luna ran on them on 28
September 2026 under the same settings. The last
rows of Table~\ref{tab:gpt6luna} give the paired differences and
Table~\ref{tab:vdem} its scores. Against JEV, with a paired country bootstrap
of \VdBootDraws{} draws, it trails
by \VdLunaCorGap{} [\VdLunaCorGapLo{}, \VdLunaCorGapHi{}] in mean per-country
correlation and by \VdLunaExactGap{} [\VdLunaExactGapLo{}, \VdLunaExactGapHi{}]
points in exact agreement. Its probabilities, read from the five alternatives
the interface returns for the answer token, put more than 0.99 on the chosen
level for \VdLunaOverNinetyNine\% of pairs, with a mean modal probability of \VdLunaPmodeMean{}
against a hit rate of \VdLunaExact\%. Its calibration error is \VdLunaEceVal{},
higher than JEV's by \VdLunaEceGap{} [\VdLunaEceGapLo{}, \VdLunaEceGapHi{}]. The
five alternatives hold almost all of the probability, so the truncation does
not explain the gap. JEV's probabilities also separate right from wrong codes better than GPT-6
Luna's and Qwen3.8-27B's (AUROC 0.692 against 0.612 and 0.673; JEV minus Luna
0.080 [0.063, 0.095], JEV minus Qwen 0.019 [0.005, 0.032]). On the survey pairs the two are level in calibration when probabilities are
averaged over both presentation orders (Appendix B); in single calls JEV's
calibration error is lower there and on the seven other labelled tasks of
Figure~\ref{fig:calib_luna}.

\subsection*{An open-weight comparator: Qwen3.8-27B}

We also performed all of the tests on an open-weight model, Qwen3.8-27B, on 30 September 2026. We ran the model through OpenRouter, pinned to a single host (Parasail, which serves the model at 8-bit precision, FP8), with reasoning off, a temperature of 0, and the twenty most probable tokens returned for each position in its reply. We sent it the same messages, byte for byte, as in the matching GPT-6 Luna request. All replies came from that host, and the only replies lost, to a host error, were those for one party duplet and one GTD incident. We tabulate the differences from JEV in Table~\ref{tab:qwen27b}, using the same conventions as in Table~\ref{tab:gpt6luna}. Qwen is level with or above JEV for four of the seven applications (tweets, sentiment, survey, and manifestos and speeches) and below JEV for three (attack type, party positions and democracy indicators). It scores above JEV for sentiment and, in the 2023 sample, for tweet relevance and problem framing, but below it for solution framing in the 2021 sample. For the seven primary tasks, it cost between 2.35 and 5.41 times as much per decision as JEV, in part because this route does not have a batch rate.

\begin{table}[htbp]
\centering

\footnotesize
\setlength{\tabcolsep}{3pt}
\begin{tabular}{>{\raggedright\arraybackslash}p{4.0cm}>{\raggedright\arraybackslash}p{3.4cm}rrll}
\toprule
Test & Statistic & Qwen & JEV & Difference & Reading \\
\midrule
Tweet relevance, 2021 & accuracy (\%) & 88.2 & 89.0 & $-$0.8 [$-$2.1, 0.4] & level \\
Tweet relevance, 2023 & accuracy (\%) & 79.6 & 76.2 & 3.4 [0.2, 6.8] & above$^{a}$ \\
Problem frame, 2021 & accuracy (\%) & 92.4 & 92.4 & 0.0 [$-$1.2, 1.2] & level \\
Problem frame, 2023 & accuracy (\%) & 69.3 & 61.4 & 7.8 [1.2, 14.5] & above \\
Solution frame, 2021 & accuracy (\%) & 81.4 & 84.7 & $-$3.4 [$-$4.4, $-$2.3] & below \\
Solution frame, 2023 & accuracy (\%) & 96.5 & 98.2 & $-$1.8 [$-$4.1, 0.0] & level \\
Tweet sentiment, few-shot & $\rho$ with experts & 0.841 & 0.740 & 0.102 [0.071, 0.131] & above \\
GTD attack type, names & accuracy & 0.680 & 0.715 & $-$0.034 [$-$0.037, $-$0.032] & below \\
GTD attack type, names & macro F1 & 0.518 & 0.555 & $-$0.038 [$-$0.049, $-$0.026] & below \\
GTD attack type, codebook & accuracy & 0.689 & 0.721 & $-$0.032 [$-$0.035, $-$0.029] & below \\
GTD attack type, codebook & macro F1 & 0.593 & 0.638 & $-$0.045 [$-$0.057, $-$0.034] & below \\
BBC conflict & conflict F1 & 0.176 & 0.321 & $-$0.145 [$-$0.278, $-$0.017] & below \\
Party positions & CHES $r$, mean of 8 years & 0.755 & 0.877 & $-$0.154 to $-$0.091 & below in 8 of 8 \\
Survey, expert pairs & F1 & 0.782 & 0.814 & $-$0.032 [$-$0.070, 0.003] & level \\
Survey, Study 1 scale & $r$ with GPT-4o & 0.942 & 0.937 & 0.004 [$-$0.000, 0.009] & level \\
Survey, calibration & error vs raters & 0.049 & 0.109 & $-$0.060 [$-$0.091, $-$0.009] & below \\
Manifestos, economic & $r$ with experts & 0.974 & 0.969 & 0.005 [$-$0.004, 0.022] & level \\
Manifestos, social & $r$ with experts & 0.940 & 0.918 & 0.022 [$-$0.005, 0.063] & level \\
Speeches, coal subsidies & $r$ with crowd & 0.905 & 0.814 & 0.092 [$-$0.001, 0.186] & level \\
Democracy indicators & mean country $r$ with V-Dem & 0.590 & 0.680 & $-$0.090 [$-$0.107, $-$0.074] & below \\
Democracy indicators & exact agreement & 0.511 & 0.543 & $-$0.032 [$-$0.044, $-$0.020] & below \\
Democracy indicators & calibration error & 0.152 & 0.057 & 0.095 [0.082, 0.107] & above \\
\bottomrule
\multicolumn{6}{l}{\scriptsize $^{a}$~Depends on the bootstrap seed: level under half of 200 other seeds.}
\end{tabular}
\normalsize
\caption{Qwen3.8-27B against JEV on every test, through OpenRouter on one host (Parasail, FP8), reasoning off, temperature 0, one pass, with the messages GPT-6 Luna received (Appendix H). Differences are Qwen3.8-27B minus JEV's primary specification, with paired bootstrap 95\% intervals; above, level and below say whether the interval lies above zero, contains it or lies below it. For calibration error lower is better. The frame rows use our reconstructed instructions. The survey calibration row averages over presentation orders and uses the top twenty reply tokens; the survey scale row is the correlation with GPT-4o's Study 1 scale.}
\label{tab:qwen27b}
\end{table}

\subsection*{Additional calibration tasks}

Four tasks in Figure~\ref{fig:calib_luna} (seven of its rows) come from outside the seven replications. We added them because each has a human label and all three models give probabilities. Three use pairwise comparison data that \citet{licht2025} reuse. In WiscAds \citep{carlson2017}, the model reads the text of two television advertisements from the 2008 US Senate elections and picks the more negative one (1,944 pairs). In Immigration \citep{carlson2017}, it picks the statement that expresses more fear or worry about the effects of immigration (3,265 pairs). In Park \citep{park2021}, it picks the paragraph from a House committee hearing that is more grandstanding and less information-seeking (2,228 pairs). The answer counted as correct is the text with the higher score on the Bradley-Terry scale estimated from the human coders' comparisons. Each model sees every pair in all four combinations of presentation order and label order, using the prompts of \citet{licht2025}.

The fourth task screens manifesto sentences for the European Union. On a random 10\% of the Manifesto Project corpus \citep{lehmann2025corpus}, 121,877 human-coded sentences, each model gives the probability that a sentence concerns the EU, and we score it against the human codes for the two EU categories (108 and 110). EU sentences are rare, so the calibration error is small for every model.

\clearpage
\section*{Appendix I: additional tables}

\begin{table}[htbp]
\centering

\small
\begin{tabular}{lrrrr}
\toprule
& & \multicolumn{2}{c}{Macro F1} & Unknown \\
\cmidrule(lr){3-4}
Model & Accuracy & Nine types & Known types & recall \\
\midrule
\multicolumn{5}{l}{\textit{Fine-tuned on GTD labels (published)}} \\
\quad ConfliBERT & \BrtCbertAcc{} & \BrtCbertFone{} & \BrtCbertKnownFone{} & \BrtCbertUnkRecall{} \\
\quad ConflLlama, 8-bit & \BrtCllQeightAcc{} & \BrtCllQeightFone{} & \BrtCllQeightKnownFone{} & \BrtCllQeightUnkRecall{} \\
\quad ConflLlama, 4-bit & \BrtCllQfourAcc{} & \BrtCllQfourFone{} & \BrtCllQfourKnownFone{} & \BrtCllQfourUnkRecall{} \\
\multicolumn{5}{l}{\textit{Prompted with the codebook}} \\
\quad JEV 1.13 & \BrtJevCbAcc{} & \BrtJevCbFone{} & \BrtJevCbKnownFone{} & \BrtJevCbUnkRecall{} \\
\quad GPT-6 Luna & \BrtLunaCbAcc{} & \BrtLunaCbFone{} & \BrtLunaCbKnownFone{} & \BrtLunaCbUnkRecall{} \\
\quad Qwen3.5-9B & \BrtQwenCbAcc{} & \BrtQwenCbFone{} & \BrtQwenCbKnownFone{} & \BrtQwenCbUnkRecall{} \\
\quad Qwen3.8-27B & 0.689 & 0.593 & 0.683 & 0.006 \\
\multicolumn{5}{l}{\textit{Prompted with label names only}} \\
\quad JEV 1.13 & \BrtJevBareAcc{} & \BrtJevBareFone{} & \BrtJevBareKnownFone{} & \BrtJevBareUnkRecall{} \\
\quad JEV 1.13, one question per type & \BrtOvrAcc{} & \BrtOvrFone{} & \BrtOvrKnownFone{} & \BrtOvrUnkRecall{} \\
\quad GPT-6 Luna & \BrtLunaBareAcc{} & \BrtLunaBareFone{} & \BrtLunaBareKnownFone{} & \BrtLunaBareUnkRecall{} \\
\quad Qwen3.5-9B & \BrtQwenBareAcc{} & \BrtQwenBareFone{} & \BrtQwenBareKnownFone{} & \BrtQwenBareUnkRecall{} \\
\quad Qwen3.8-27B & 0.680 & 0.518 & 0.599 & 0.000 \\
\quad Qwen 2.5 14B (published) & \BrtQwenTwoAcc{} & \BrtQwenTwoFone{} & \BrtQwenTwoKnownFone{} & \BrtQwenTwoUnkRecall{} \\
\quad Llama 3.1 8B (published) & \BrtLlamaAcc{} & \BrtLlamaFone{} & \BrtLlamaKnownFone{} & \BrtLlamaUnkRecall{} \\
\quad Gemma 2 9B (published) & \BrtGemmaAcc{} & \BrtGemmaFone{} & \BrtGemmaKnownFone{} & \BrtGemmaUnkRecall{} \\
\bottomrule
\end{tabular}
\normalsize
\caption{Attack type of \BrtN{} GTD incidents, replicating \citet{brandt2026}. Known types: macro F1 over the eight types, on the \BrtNKnown{} incidents not coded Unknown. Unknown recall: share of the \BrtNUnknown{} Unknown incidents labelled Unknown. GPT-6 Luna was sent Qwen3.5-9B's prompts, and the open-weight Qwen3.8-27B received GPT-6 Luna's requests (Appendix H).}
\label{tab:brandt}
\end{table}

\begin{table}[htbp]
\centering

\small
\setlength{\tabcolsep}{4pt}
\begin{tabular}{lrrrrrr}
\toprule
Coder & Correlation & Difference & Exact & Changed & Within one & Missing \\
\midrule
JEV, most probable level & 0.68 [0.66, 0.70] & $-$0.16 & 54 & 36 & 92 & 0 \\
JEV, expected level & 0.75 [0.73, 0.76] & $-$0.20 &  &  &  & 0 \\
GPT-6 Luna & 0.63 [0.61, 0.64] & $-$0.09 & 52 & 38 & 91 & 0 \\
Qwen3.8-27B & 0.59 [0.57, 0.61] & $-$0.17 & 51 & 35 & 89 & 0 \\
Qwen3.8-27B, expected level & 0.67 [0.66, 0.69] & $-$0.18 &  &  &  & 0 \\ %
GPT-4o & 0.63 [0.62, 0.65] & $-$0.28 & 49 & 36 & 90 & 5 \\
Llama-3.1 70B & 0.50 [0.49, 0.52] & $+$0.50 & 46 & 29 & 82 & 129 \\
GPT-4o and Llama, averaged & 0.64 [0.62, 0.65] & $+$0.11 &  &  &  & 134 \\
\bottomrule
\end{tabular}
\normalsize
\caption{V-Dem indicators coded from the country name alone, replicating \citet{weidmann2026}: 9,041 country-indicator pairs (53 ordinal indicators, 171 countries, 2023), scored against V-Dem v14. Correlation is the mean over countries of the correlation across a country's indicators, with a 95\% country-bootstrap interval. Difference is the mean of the model's code minus V-Dem's (below zero is pessimistic). Exact, Changed (the 877 pairs whose value changed from 2022) and Within one are percentages of pairs; Missing counts pairs without a code. GPT-4o and Llama-3.1 rows rescore the published codes (the article reports 0.64 for GPT-4o). GPT-6 Luna and Qwen3.8-27B received the published prompt (Appendix H).}
\label{tab:vdem}
\end{table}

\begin{table}[htbp]
\centering

\footnotesize
\setlength{\tabcolsep}{4pt}
\textit{a. GPT-3.5, JEV and GPT-6 Luna against three benchmarks}\par\smallskip
\begin{tabular}{lrrrrrrrrr}
\toprule
& \multicolumn{3}{c}{CHES experts} & \multicolumn{3}{c}{TEV voters} & \multicolumn{3}{c}{CMP manifestos} \\
\cmidrule(lr){2-4}\cmidrule(lr){5-7}\cmidrule(lr){8-10}
Year & GPT-3.5 & JEV & \shortstack{GPT-6\\Luna} & GPT-3.5 & JEV & \shortstack{GPT-6\\Luna} & GPT-3.5 & JEV & \shortstack{GPT-6\\Luna} \\
\midrule
1979 & --- & --- & --- & 0.75 & 0.92 {\scriptsize [0.79, 0.97]} & 0.93 & 0.49 & 0.57 {\scriptsize [0.44, 0.69]} & 0.51 \\
1984 & 0.77 & 0.88 {\scriptsize [0.81, 0.92]} & 0.83 & 0.78 & 0.92 {\scriptsize [0.84, 0.96]} & 0.89 & 0.53 & 0.59 {\scriptsize [0.46, 0.69]} & 0.56 \\
1989 & 0.77 & 0.88 {\scriptsize [0.81, 0.92]} & 0.84 & 0.74 & 0.82 {\scriptsize [0.65, 0.91]} & 0.79 & 0.60 & 0.59 {\scriptsize [0.46, 0.69]} & 0.59 \\
1994 & 0.81 & 0.89 {\scriptsize [0.83, 0.93]} & 0.84 & 0.84 & 0.89 {\scriptsize [0.79, 0.94]} & 0.83 & 0.44 & 0.52 {\scriptsize [0.43, 0.60]} & 0.50 \\
1999 & 0.75 & 0.88 {\scriptsize [0.83, 0.91]} & 0.84 & 0.77 & 0.88 {\scriptsize [0.81, 0.93]} & 0.85 & 0.44 & 0.49 {\scriptsize [0.39, 0.57]} & 0.48 \\
2004 & 0.76 & 0.86 {\scriptsize [0.81, 0.89]} & 0.81 & 0.68 & 0.81 {\scriptsize [0.72, 0.87]} & 0.76 & 0.53 & 0.55 {\scriptsize [0.46, 0.64]} & 0.59 \\
2009 & 0.78 & 0.88 {\scriptsize [0.84, 0.91]} & 0.86 & 0.58 & 0.64 {\scriptsize [0.48, 0.76]} & 0.60 & 0.44 & 0.45 {\scriptsize [0.34, 0.55]} & 0.52 \\
2014 & 0.76 & 0.87 {\scriptsize [0.84, 0.90]} & 0.85 & 0.59 & 0.84 {\scriptsize [0.74, 0.90]} & 0.80 & 0.55 & 0.55 {\scriptsize [0.46, 0.63]} & 0.58 \\
2019 & 0.78 & 0.89 {\scriptsize [0.86, 0.92]} & 0.87 & --- & --- & --- & 0.59 & 0.62 {\scriptsize [0.54, 0.69]} & 0.65 \\
\bottomrule
\end{tabular}

\medskip
\textit{b. Qwen3.8-27B, and the CSES voter benchmark for all three models}\par\smallskip
\begin{tabular}{lrrrrrr}
\toprule
& \multicolumn{3}{c}{Qwen3.8-27B} & \multicolumn{3}{c}{CSES voters} \\
\cmidrule(lr){2-4}\cmidrule(lr){5-7}
Year & CHES & TEV & CMP & JEV & \shortstack{GPT-6\\Luna} & \shortstack{Qwen\\27B} \\
\midrule
1979 & --- & 0.86 & 0.46 & --- & --- & --- \\
1984 & 0.79 & 0.81 & 0.51 & --- & --- & --- \\
1989 & 0.76 & 0.78 & 0.50 & --- & --- & --- \\
1994 & 0.79 & 0.82 & 0.37 & --- & --- & --- \\
1999 & 0.78 & 0.84 & 0.38 & 0.84 & 0.78 & 0.66 \\
2004 & 0.70 & 0.72 & 0.48 & 0.80 & 0.73 & 0.67 \\
2009 & 0.74 & 0.57 & 0.42 & 0.82 & 0.77 & 0.67 \\
2014 & 0.73 & 0.65 & 0.54 & 0.76 & 0.70 & 0.58 \\
2019 & 0.75 & --- & 0.57 & --- & --- & --- \\
\bottomrule
\end{tabular}
\normalsize
\caption{European party positions, replicating \citet{dileo2025}: Pearson correlation of Bradley-Terry scores with each benchmark by year. CHES: Chapel Hill Expert Survey (before 1999, the Ray--Marks--Steenbergen values); TEV: True European Voter surveys; CMP: Manifesto Project scores. Their GPT-3.5 figures, the modal answer over seven runs, are transcribed from their Figure 1; dashes mark panels they leave empty. JEV's are a single pass with 95\% Fisher intervals; GPT-6 Luna and, in panel b, Qwen3.8-27B are single passes without intervals (Appendix H). Panel b adds voters' placements in the Comparative Study of Electoral Systems (CSES), available in their files for 1999 to 2014.}
\label{tab:dileo}
\end{table}

\begin{table}[htbp]
\centering

\small
\begin{tabular}{lrrr}
\toprule
& \multicolumn{2}{c}{UK manifestos (experts)} & EU speeches (crowd) \\
\cmidrule(lr){2-3}\cmidrule(lr){4-4}
& Economic & Social & Subsidies \\
\midrule
JEV & 0.97 & 0.92 & 0.81 \\
JEV, relevance-weighted & 0.97 & 0.92 & 0.86 \\
GPT-4o (published) & 0.98 & 0.94 & 0.93 \\
GPT-6 Luna & 0.98 & 0.95 & 0.95 \\
Qwen3.8-27B & 0.97 & 0.94 & 0.90 \\
Other published models & 0.95 to 0.98 & 0.88 to 0.95 & $-$0.52 to 0.86 \\
\midrule
JEV $-$ GPT-4o & $-$0.01 {\scriptsize [$-$0.03, 0.00]} & $-$0.02 {\scriptsize [$-$0.05, 0.00]} & $-$0.12 {\scriptsize [$-$0.20, $-$0.04]} \\
JEV, weighted $-$ GPT-4o & $-$0.01 {\scriptsize [$-$0.02, 0.00]} & $-$0.02 {\scriptsize [$-$0.05, 0.00]} & $-$0.07 {\scriptsize [$-$0.15, +0.02]} \\
JEV $-$ GPT-6 Luna & $-$0.01 {\scriptsize [$-$0.02, $-$0.00]} & $-$0.03 {\scriptsize [$-$0.08, $-$0.01]} & $-$0.14 {\scriptsize [$-$0.21, $-$0.07]} \\
JEV $-$ Qwen3.8-27B & $-$0.01 {\scriptsize [$-$0.02, 0.00]} & $-$0.02 {\scriptsize [$-$0.06, +0.01]} & $-$0.09 {\scriptsize [$-$0.19, 0.00]} \\
\midrule
Documents & 18 & 18 & 36 \\
Sentences & 18,263 & 18,263 & 429 \\
\bottomrule
\end{tabular}
\normalsize
\caption{Party manifestos and legislative speeches, replicating Le Mens and
Gallego (2025): Pearson correlation of document positions with the benchmark.
Positions average the sentence scores of the sentences judged relevant; for
JEV, those with a relevance probability of at least 0.5, and in the weighted
row every sentence, weighted by that probability. Published figures are
recomputed from their archive. Gaps carry 95\% intervals from a paired
bootstrap over documents. GPT-6 Luna and the open-weight Qwen3.8-27B received the published prompts, scale and NA option, and average the sentences they did not mark NA; their gaps with JEV use the 35 speeches on which JEV judged at least one sentence relevant.}
\label{tab:lemens}
\end{table}

\end{document}